\documentclass{article}

\usepackage{arxiv}

\usepackage[utf8]{inputenc} 
\usepackage[T1]{fontenc}    
\usepackage{hyperref}       
\usepackage{url}            
\usepackage{booktabs}       
\usepackage{amsfonts}       
\usepackage{nicefrac}       
\usepackage{microtype}      

\usepackage{lipsum}         
\usepackage{graphicx}
\usepackage{natbib}
\usepackage{doi}

\usepackage{amsmath,amsfonts,bm}

\def\eqref#1{equation~\ref{#1}}

\def\1{\bm{1}}

\DeclareMathAlphabet{\mathsfit}{\encodingdefault}{\sfdefault}{m}{sl}
\SetMathAlphabet{\mathsfit}{bold}{\encodingdefault}{\sfdefault}{bx}{n}

\usepackage{hyperref}
\usepackage{url}
\usepackage{colortbl}
\usepackage{xcolor}
\usepackage[dvipsnames]{xcolor}
\usepackage{amsmath}
\usepackage{amsmath,amsfonts}
\usepackage{algorithmic}
\usepackage{algorithm}
\usepackage{wrapfig}
\usepackage{siunitx}

\usepackage{algorithm}
\usepackage{dsfont}

\usepackage[textsize=tiny]{todonotes}
\usepackage{enumitem}

\usepackage{amsmath,amsfonts}
\usepackage{algorithmic}
\usepackage{algorithm}
\usepackage{array}
\usepackage[caption=false,font=normalsize,labelfont=sf,textfont=sf]{subfig}
\usepackage{textcomp}
\usepackage{stfloats}
\usepackage{url}
\usepackage{verbatim}
\usepackage{graphicx}

\usepackage{hyperref}
\usepackage{url}
\usepackage{colortbl}
\usepackage{xcolor}
\usepackage[dvipsnames]{xcolor}
\usepackage{amsmath}
\usepackage{wrapfig}
\usepackage{siunitx}
\usepackage{algorithm}
\usepackage{dsfont}

\usepackage[textsize=tiny]{todonotes}
\usepackage{enumitem}

\title{"\textit{What Is The Performance Ceiling of My Classifier}?"\\ Utilizing Category-Wise Influence Functions for Pareto Frontier Analysis}

\date{}

\usepackage{authblk}

\author[1]{%
    Shahriar Kabir Nahin
}
\author[2]{%
    Wenxiao Xiao%
}

\author[2]{%
     Joshua Liu%
}
\author[1]{%
    Anshuman Chhabra%
}

\author[2]{%
     Hongfu Liu%
}

\affil[1]{Bellini College of AI, Cybersecurity, and Computing, University of South Florida\thanks{Email: \texttt{\{shahriarkabir, anshumanc\}@usf.edu}}}
\affil[2]{Michtom School of Computer Science, Brandeis University\thanks{Email: \texttt{\{wenxiaoxiao, joshliu, hongfuliu\}@brandeis.edu}}}

\renewcommand{\shorttitle}{"\textit{What Is The Performance Ceiling of My Classifier}?"}

\begin{document}

\maketitle

\begin{abstract}
\looseness-1 Data-centric learning seeks to improve model performance from the perspective of data quality, and has been drawing increasing attention in the machine learning community. Among its key tools, influence functions provide a powerful framework to quantify the impact of individual training samples on model predictions, enabling practitioners to identify detrimental samples and retrain models on a cleaner dataset for improved performance. 
However, most existing work focuses on the question \textit{``what data benefits the learning model?"} In this paper, we take a step further and investigate a more fundamental question: \textit{``what is the performance ceiling of the learning model?"} Unlike prior studies that primarily measure improvement through overall accuracy, we emphasize category-wise accuracy and aim for Pareto improvements, ensuring that every class benefits, rather than allowing tradeoffs where some classes improve at the expense of others. To address this challenge, we propose category-wise influence functions and introduce an influence vector that quantifies the impact of each training sample across all categories. Leveraging these influence vectors, we develop a principled criterion to determine whether a model can still be improved, and further design a linear programming–based sample reweighting framework to achieve Pareto performance improvements. Through extensive experiments on synthetic datasets, vision, and text benchmarks, we demonstrate the effectiveness of our approach in estimating and achieving a model's performance improvement across multiple categories of interest. 

\end{abstract}

\section{Introduction}\label{introduction}
Data-centric learning has recently emerged as a central research topic in the machine learning community \citep{feldman2020neural, chhabra2024what, richardson2023add}, shifting attention from purely algorithmic model design towards improving the quality of training data. Unlike conventional preprocessing techniques such as normalization or outlier removal, that operate independently of the learning algorithm, data-centric learning is tightly coupled with the downstream model. Its primary goal is to assess whether each training sample is beneficial or detrimental with respect to a specific learning objective, thereby guiding principled data curation and optimization.

Sample influence estimation is a fundamental task in data-centric learning. The seminal work of \citet{koh2017understanding} introduced \textit{influence functions}, a technique from robust statistics \citep{hampel1974influence, cook1982residuals} that enables approximate influence estimation without retraining. This approach allows efficient estimation of sample influence, inspiring a series of follow-up studies and variants on its applications on deep models~\citep{schioppa2022scaling,chhabra2025outlier,kwon2023datainf,wang2024data}.

\looseness-1Broadly speaking, sample influence estimation seeks to answer the question: “\textit{which training samples help the model, and which ones harm it}~\citep{chhabra2024what}?” In practice, samples identified as detrimental are removed, and the model is retrained on the refined dataset, often resulting in measurable performance gains. This naturally raises deeper questions: \textit{can the model’s performance be improved even further by iteratively repeating this process? And if so, what is the ultimate performance ceiling of the learning model?}

In this paper, we focus on the multi-class classification setting: given a training set, a validation set, and a classification model, we aim to answer two key questions from the data perspective: (1) has the classifier already reached its maximum potential performance, and (2) if not, how can its performance be further improved? Importantly, the notion of “improvement” in our work does not simply refer to an increase in overall accuracy. Instead, we adopt a \textit{Pareto improvement} perspective, seeking performance gains for the target class while preserving performance on other classes, thereby avoiding trade-off scenarios in which improvements for some classes come at the expense of others. We summarize our contributions as follows:

\begin{itemize}[wide=10pt, leftmargin=*, nosep]
\item We tackle a fundamental yet largely overlooked question: “\textit{What is the performance ceiling of a classifier?}” Specifically, we determine whether a given classifier has already reached its maximum potential performance and, if not, how it can be further improved.\vspace{1mm}
\item We introduce category-wise\footnote{We use the terms \textit{category} and \textit{class} interchangeably throughout the paper.} influence functions to assess the model’s Pareto frontier on each category for the analysis on performance ceiling. Leveraging these influence scores, we further propose a linear programming–based sample reweighting framework to achieve Pareto performance improvements across classes of interest.\vspace{1mm}
\item We validate our category-wise influence functions on both synthetic and benchmark datasets and present detailed case studies showing how to determine whether a classifier has reached its performance ceiling using our linear programming–based sample reweighting framework to enable targeted model refinement.
\end{itemize}

\section{Related Work}

\looseness-1 \textbf{Sample Influence Estimation.} Influence functions comprise a set of methods from robust statistics \citep{hampel1974influence, cook1982residuals} that can be used to approximately estimate influence without requiring retraining, i.e., they can help create a conceptual link that traces model performance to samples in the training set. For gradient-based models trained using empirical risk minimization, the seminal work by \citet{koh2017understanding} utilizes a Taylor-series approximation and LiSSA optimization \citep{agarwal2017second} to compute sample influences and relies on the Hessian matrix. Follow-up works such as Representer Point \citep{yeh2018representer} and Hydra \citep{chen2021hydra} improve influence estimation performance for deep learning models. More recently, efficient influence estimation methods such as DataInf \citep{kwon2023datainf}, Arnoldi iteration \citep{schioppa2022scaling}, and Kronecker-factored approximation curvature \citep{grosse2023studying} have been proposed which can even be employed for larger models. Some other approaches directly utilize the gradient space to measure influence \citep{pruthi2020estimating, charpiat2019input}, while others use some ensemble methods~\citep{bae2024training,kim2024gex, dai2023training}.
Recent work has also found that \textit{self-influence} only on the training set can be a useful measure for detecting sample influence \citep{bejan2023make, thakkar2023self}. Influence functions have been widely used in the community for a number of data-centric applications \citep{feldman2020neural, chhabra2024what, richardson2023add}, but the focus of these works has predominantly been on using the overall accuracy of the model as a proxy for performance measurement. This contrasts with the main motivation in our paper, where we seek to study how different samples in the training set influence different categories/classes of the data. Such class-wise tradeoff analysis based on the Pareto frontier is of paramount importance in multiple applications, such as category-aware domain adaptation \citep{wenxiao_category} and fair classification \citep{fair_pareto, martinez2020minimax}.\vspace{1mm}

\textbf{Pareto frontier Analysis.}
\looseness-1Pareto frontier analysis is widely employed in many domains, where multiple objectives need to be optimized simultaneously, necessitating solutions that can effectively measure the tradeoffs between each of the objectives. For instance, \cite{NEURIPS2023_960573a3} analyzes Pareto tradeoffs across resources for model training including data, model architectures, and computation. \cite{NEURIPS2019_685bfde0} introduces Pareto multi-task learning, where multiple Pareto-optimal solutions are generated efficiently to help practitioners choose the best one according to their tradeoff preferences. Other work, such as the Iterated Pareto Referent Optimization method proposed by \cite{10.5555/3709347.3743813} for multi-objective reinforcement learning, decomposes the search for optimal paths into a sequence of single objectives. Similarly, \cite{NEURIPS2023_32285dd1} handles distributional pareto-optimal policies in reinforcement learning under uncertainty. Pareto-optimal tradeoffs have also been studied in other problem domains, such as neural architecture search \citep{JMLR:v20:18-598}. As is evident, none of these works investigate category-wise tradeoffs during training and analysis of the Pareto-front for assessing the performance ceiling of a given classifier, unlike our work.\vspace{1.5mm}


\looseness-1\textbf{Other Data-Centric Learning.} Many works in data-centric learning study research questions beyond Pareto frontier analysis and category-wise influence estimation. \textit{Datamodels} \citep{ilyas2022datamodels} estimate training sample contributions as well, but only for one test sample at a time. Other approaches such as \citep{jain2022efficient, paul2021deep, killamsetty2021retrieve} aim to accelerate deep learning training time via subset/coreset selection. Data pruning, augmentation, and relabeling approaches \citep{yang2022dataset, tan2024data, kong2021resolving, chhabra2022fair, richardson2023add} and model pruning approaches \citep{lyu2023deeper} based on influence analysis have also been proposed. Another related area of research is \textit{active learning} \citep{cohn1996active}, which seeks to iteratively identify optimal samples to annotate given a large unlabeled training data pool \citep{isal, nguyen2022measure, wei2015submodularity}. 

\vspace{-2mm}
\section{Proposed Approach}

\subsection{Preliminaries}

Let $T$$=$$\{z_i\}_{i=1}^n$ be a training set, where $z_i$$=$$(x_i, y_i)$ includes the input space sample features $x_i$ and output space label $y_i$. A classifier trained on the empirical loss $\ell$ can be written as: $\Hat{\theta}$$=$$\arg\min_{\theta \in \Theta} \frac{1}{n} \sum_{i=1}^n \ell(z_i; \theta)$. Influence functions~\citep{koh2017understanding} constitute methods from robust statistics~\citep{hampel1974influence,cook1982residuals,martin1986influence} that can help measure the effect of changing an infinitesimal weight of training samples on the model utility/performance. Downweighting a training sample $z_j$ by a very small fraction $\epsilon$ leads to a model parameter: $\Hat{\theta}(z_j; -\epsilon) = \arg\min_{\theta \in \Theta} \frac{1}{n} (\sum_{i=1}^n \ell(z_i; \theta)$$-$$\epsilon \ell(z_j; \theta))$. By evaluating the limit as $\epsilon$ approaches 1, we can estimate the \textit{influence score} associated with the removal of $z_j$ from the training set in terms of loss on the validation set $V$, without undertaking any computationally expensive leave-one-out re-training as:
\begin{equation}
\label{eq:influence}
\begin{aligned}
\mathcal{I}^{\Hat{\theta}}(z_j, V) = \sum\nolimits_{z \in V}\nabla_{\Hat{\theta}}\ell(z; \hat{\theta})^{\top}\mathbf{H}^{-1}_{\Hat{\theta}}\nabla_{\hat{\theta}}\ell(z_j; \Hat{\theta}).
\end{aligned}
\end{equation}
where $\nabla_{\Hat{\theta}}\ell(z_j; \Hat{\theta})$ is the gradient of sample $z_j$ to model parameters, and $\mathbf{H}_{\Hat{\theta}}$$=$$\sum_{i=1}^n \nabla_{\Hat{\theta}}^2 \ell(z_i; \Hat{\theta})$ denotes the Hessian. 

\looseness-1Higher values of $\mathcal{I}^{\Hat{\theta}}(z_j,V)$ indicate a more positively influential sample (i.e., one that decreases the overall classification loss) and conversely, lower values correspond to a more negatively influential sample. Also note that while influence functions have demonstrated their benefits in deep learning non-convex models, such as on BERT~\citep{han2020explaining}, ResNets~\citep{isal, yang2022dataset}, and CNNs~\citep{koh2017understanding, schioppa2022scaling}, there is ongoing research that studies their suitability to these models~\citep{bae2022if, basu2020influence, epifano2023revisiting, schioppa2024theoretical}. While this research question regarding applicability is not the focus of our paper we employ influence function formulations that have been shown to work well for deep models \citep{grosse2023studying, kwon2023datainf}.

\subsection{Research Question}


While measuring the influence of training samples on the predictive performance of a model can serve as a powerful tool for numerous data-centric learning applications, prior work \citep{koh2017understanding, chhabra2024what, han2020explaining, kwon2023datainf, schioppa2024theoretical, isal, yang2022dataset, chhabra2025outlier} has undertaken this analysis using only the \textit{overall accuracy} of the model as an indicator for performance. However, this is a restrictive scenario, and it can be beneficial for users/developers to consider a category-wise analysis for fine-grained understanding model's performance. 

\looseness-1The basis for category-wise analysis stems from the fact that different training samples can lead the model to learn different predictive patterns, and hence, they can impact their own class and other classes in varying ways. Following this rationale, it should be possible to ascertain how different samples in the training set impact different categories of the dataset and therefore, utilize this information as a model developer to make relevant tradeoffs that are desirable for the given application at hand. These tradeoffs appear in numerous learning problems-- such as \textit{fair classification}~\citep{fair_pareto, martinez2020minimax}, where performance of multiple different categories need to be maximized jointly and \textit{category-aware active domain adaptation}~\citep{wenxiao_category}, where performance impacts across different categories/classes need to be individually identified, among several others. In contrast to past work which focuses solely on overall class accuracy, this forms the main motivation for our research focus in this work, where we seek to study how performance can be improved for certain classes by potentially sacrificing performance of others. The space of solutions that describe these different tradeoffs between categories is also known as the \textit{Pareto frontier} \citep{lotov2008visualizing}. 

\looseness-1In this work, we consider whether a given classifier has already reached its maximum potential performance, and if not, how it can be further improved. We will now discuss our proposed methods for obtaining the Pareto frontier (i.e., the classifier's performance ceiling).

\subsection{Category-wise Influence Vector Estimation}
\looseness-1 We now discuss our proposed approach for obtaining the Pareto frontier for the classifier across different categories. Analytically, the Pareto frontier for a training sample can be described using an \textit{influence vector} ${P}(z) \in \mathbb{R}^K$ of length $K$ (given $K$ classes/categories) where each cell ${P}^k(z)$ for $k \in [K]$ is the impact to the category $k$ when the sample $z$ is removed from the training set. Next, we will develop category-wise influence scores for obtaining this Pareto frontier vector.

\looseness-1Let the subset of samples of set $S$ that belong to class $k$ be denoted as $S^k$. We can then measure the influence score for training sample $z$ as ${P}^k(z) = \mathcal{I}^{\Hat{\theta}}(z, S^k)$ for all $k \in [K]$ using Eq. (\ref{eq:influence}). It is important to note that the category-wise influence vector ${P}(z)$ is a useful solution aimed at answering the research question (\textit{what is the classifier's performance ceiling?}) we had formulated above. More specifically, the influence vector allows users/developers to easily gauge the classifier's performance ceiling-- if all values of ${P}(z) > 0$, the sample $z$ is \textit{beneficial} to all categories and if all values of ${P}(z) \leq 0$, the sample $z$ is \textit{detrimental} to all categories. The third case is when ${P}(z)$ takes on \textit{mixed} values that are both positive and negative.

\begin{wrapfigure}{r}{0.4\textwidth}
    \centering
    \includegraphics[width=\linewidth]{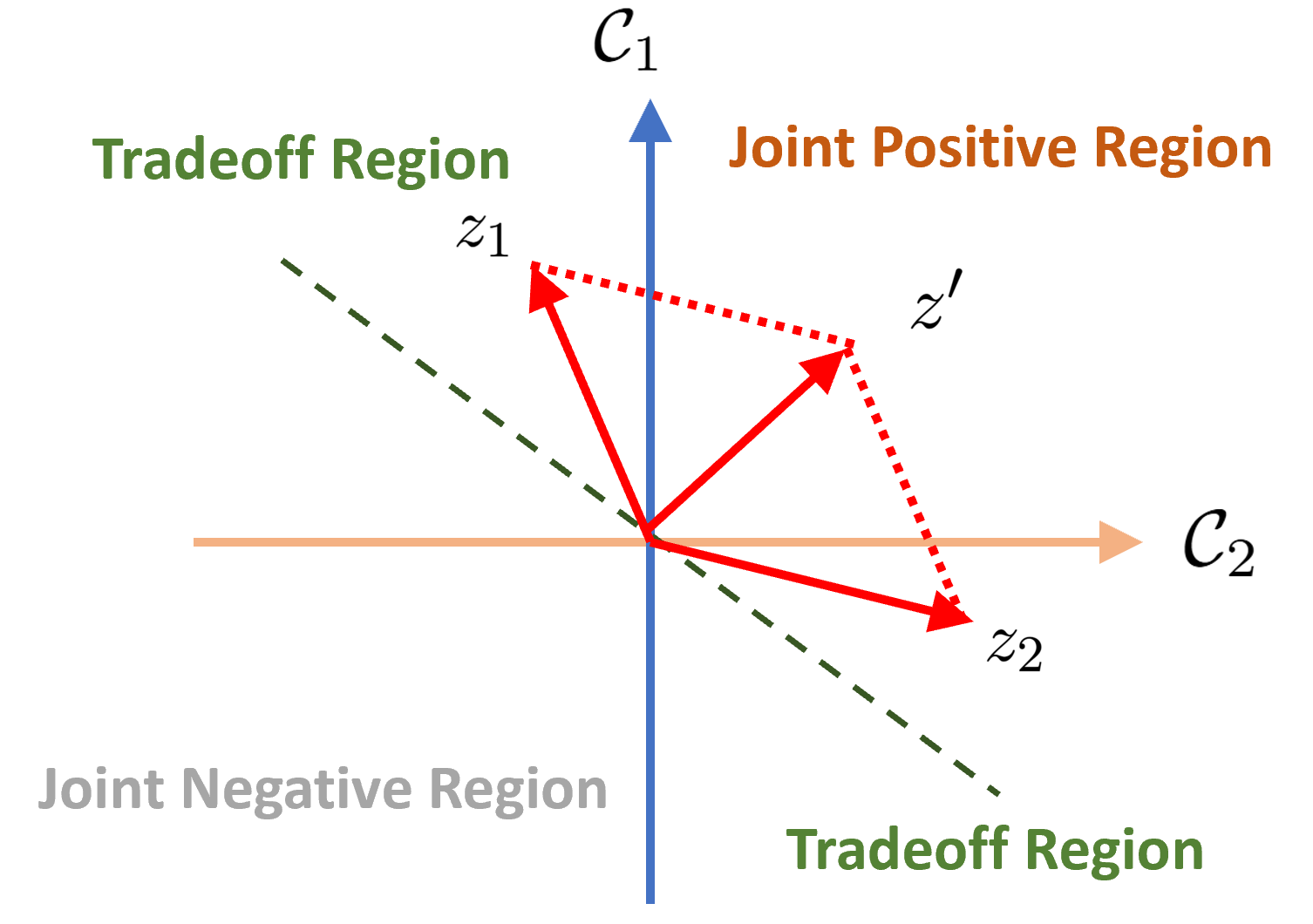}
    \vspace{-2mm}
    \caption{Influence space for two categories.}
    \label{fig:tradeoff}
    \vspace{-5mm}
\end{wrapfigure}

\looseness-1Without loss of generalizability, consider the case scenario with two categories: $\mathcal{C}_1$ and $\mathcal{C}_2$. Given a model, we can calculate the influence vector for all training samples, and visualize the influence vector in Fig.~\ref{fig:tradeoff}. Training samples located in the joint negative region are expected to hinder performance in both categories; thus, their removal could lead to simultaneous improvements in both aspects. Conversely, samples in the joint positive region can be further leveraged or retrained to enhance performance. These observations imply that the presence of samples in either the joint negative or positive regions indicates room for Pareto improvement, suggesting that the current model has not yet reached the Pareto frontier. However, an important question arises: \textit{if no samples exist in the joint negative or positive regions, does this imply that the Pareto frontier has been achieved?} 



\looseness-1The answer to the above question is no, indicating that the current model can still be improved. At the individual sample level, if all samples may be located within tradeoff regions, involving or removing any sample will lead the tradeoff effect. However, when samples are considered collectively as a set, combining certain samples can yield a new sample that falls into the joint positive region. For example, as shown by the red arrows $z_1$ and $z_2$ in Fig.~\ref{fig:tradeoff}, these two training samples lie in different tradeoff regions. Yet, when combined, the new sample becomes jointly beneficial to both categories. This observation suggests that a reweighting strategy \textit{can} be employed to modify the training set for Pareto improvements. Unlike approaches that simply decide whether to include or exclude individual samples, reweighting considers combinations of samples, providing a more general and flexible mechanism for optimization. Based on this insight, we can establish a new condition for achieving the Pareto frontier: all samples are close to the line $y=-x$ (the dashed green line). In the following, we will provide a reweighting framework based on the influence vector to achieve Pareto improvement.

\subsection{Improving Pareto Performance Using Influence Vectors}

\looseness-1We now aim to utilize our category-aware influence vector $P(z), \forall z\in T$ to improve the performance of a given classifier. More specifically, we utilize the category-wise influence vector to obtain per-sample weights for training losses that improve performance on a target subset of categories while controlling degradation on the remaining categories, via \textit{linear programming} (LP) \citep{dantzig2002linear}. 
Furthermore, while influence scores are useful estimators for tuning category-wise performance, they cannot be utilized to obtain class-wise performance thresholds (equivalently, \textit{slack variables} in the linear program). Hence, we utilize a \textit{genetic algorithm} (GA) \citep{forrest1996genetic} to help identify these class-specific slack variables and ensure that the weighted LP obtains highly optimized solutions.
We thus propose our approach \textsc{Pareto-LP-GA}, which is an influence vector-guided linear programming approach for training sample weight optimization combined with a GA search. 

\looseness-1We apply this weighted model in the context of two different settings: (a) \textit{Direct Improvement (DI)}: this refers to improving specific categories in a particular epoch as desired by the model developer where target categories are selected by the developer based on current per-class accuracy observations; and (b) \textit{Course Correction (CC)}: the developer while training the model observes accuracy drops in a certain epoch, and decides to modify the training trajectory for the detrimental epoch identified. A detrimental epoch means one after which the accuracy of some classes decreases significantly, indicating potential Pareto-optimal class/category tradeoffs.

\begin{algorithm}[t]
\fontsize{9}{9}\selectfont
\caption{\textsc{Pareto-LP-GA}}
\label{alg:influence-ga}
\begin{algorithmic}[1]
\REQUIRE Training set $T$, model params $\hat{\theta}^e$ for epoch $e$, influence vector $P(z), \forall z \in T$, target set $\mathcal{C}_{\text{target}}$, GA iterations $G$
\ENSURE Optimized per-sample weights ${w^*}$, optimized class-wise performance thresholds ${\alpha^*}$
\STATE \textbf{initialize} population $\alpha^0 = \{\alpha^0_k\}_{k\in [K]}$ randomly
\FOR{$g \in [G]$}
    \FOR{candidate threshold set $\alpha^g$ in current population}
        \STATE \textbf{solve} the following LP to obtain $w$:
        \vspace{-0.5em}
        \begin{align*}
        \max_{w} \sum\nolimits_{k \in \mathcal{C}_{\text{target}}} \sum\nolimits_{z_i\in T} w_i P^k(z_i) \quad
        \\ \text{s.t.} \sum\nolimits_{z_i\in T} w_i P^k(z_i) \geq \alpha^g_k \sum\nolimits_{z_i \in T} P^k(z_i), \forall k \in [K]
        \end{align*}
        \vspace{-0.5em}
        \STATE $\hat{\theta}^{e+1} \gets \textsc{TrainOneEpoch}(\hat{\theta}^e, T, w)$
        \STATE \textbf{compute} relative change in performance $\Delta_k^{e+1}$ from $\hat{\theta}^e$ to $\hat{\theta}^{e+1}$ for $k \in [K]$
        \STATE \textbf{compute} fitness $F(\alpha^g)$ for current candidate threshold set as follows:
        \vspace{-0.5em}
        {\small
        \begin{align*}
        F(\alpha^g) = \frac{1}{|\mathcal{C}_{\text{target}}|} \sum_{k \in \mathcal{C}_{\text{target}}} \mathds{1}_{[\Delta_k^{e+1} \leq 0]} (-\infty) \\+ \frac{1}{|\mathcal{C} \setminus \mathcal{C}_{\text{target}}|} \sum_{k \notin \mathcal{C}_{\text{target}}} \mathds{1}_{[\Delta_k^{e+1} < 0]} (\Delta_k^{e+1})
        \end{align*}}
        \vspace{-3mm}
    \ENDFOR
    \STATE \textbf{apply} selection, crossover, and mutation operations on population
    \STATE \textbf{store} $\alpha^*$ and $w^*$ that maximizes fitness so far
\ENDFOR
\STATE \textbf{return} ${w^*}, {\alpha^*}$
\end{algorithmic}
\end{algorithm}

\looseness-1The \textsc{Pareto-LP-GA} procedure is provided in Algorithm \ref{alg:influence-ga}. Denoting the full set of $K$ classes as $\mathcal{C}$, let $\mathcal{C}_{\text{target}}$ be the target subset of categories we wish to improve the performance for, while ensuring minimal performance degradation in other classes $\mathcal{C} \setminus \mathcal{C}_{\text{target}}$. Algorithm \ref{alg:influence-ga} takes in as input the training set $T$, the model parameters trained until a certain epoch $e$, our category-aware influence vector $P(z)$, the target classes $\mathcal{C}_{\text{target}}$ to improve performance for, the total iterations $G$. First, we initialize the population variable that controls for class-wise performance threshold along the Pareto frontier randomly, denoted as $\alpha_k, \forall k \in [K]$. Then, for each iteration of the genetic algorithm (GA), we solve a linear program (Line 4) that seeks to optimize for the per-sample weights by ensuring estimated performance via the category-aware influence vector is maximized on target classes while ensuring class-wise performance is above each category/class threshold ($\alpha_k$). Subsequently, we train the model for the next epoch ($e+1$) by applying the current optimized weight set (Line 5). 

\looseness-1 We then measure relative change in performance ($\Delta_k^{e+1}$) between epochs $e \rightarrow e+1$ for \textit{Direct Improvement} (for \textit{Course Correction} this performance change $\Delta_k^{e+1}$ is instead calculated between the original epoch $e+1$ and the newly weighted epoch $e+1$; the rest of the procedure remains identical). Now, while we have optimized the weight set, we still need to obtain the optimal class-wise thresholds via the GA search. Hence, we formulate the fitness function (Line 7) such that if performance for $\mathcal{C}_{\text{target}}$ decreases at all (i.e., $\Delta_k^{e+1} \leq 0$) the fitness is set to a large-magnitude negative value (denoted as $-\infty$). Moreover, for non-target classes $\mathcal{C} \setminus \mathcal{C}_{\text{target}}$, if performance decreases, the fitness score reflects the degree of degradation. Thus, the GA $\alpha$ search also optimizes for performance improvement along desired target classes while ensuring minimal performance reduction across non-target classes. Post this step, we apply the standard GA operations (selection, crossover, mutation, etc.) on the population. Eventually, the algorithm return the optimized weight set $w^* = [w_1^*, \ldots, w_n^*]$ for the training set, that will be applied to the loss computation during next epoch training. Now, we describe the experimental setup for our proposed approach and present the results in the subsequent sections.

\begin{figure*}[t]
    \centering
    \includegraphics[width=0.98\linewidth]{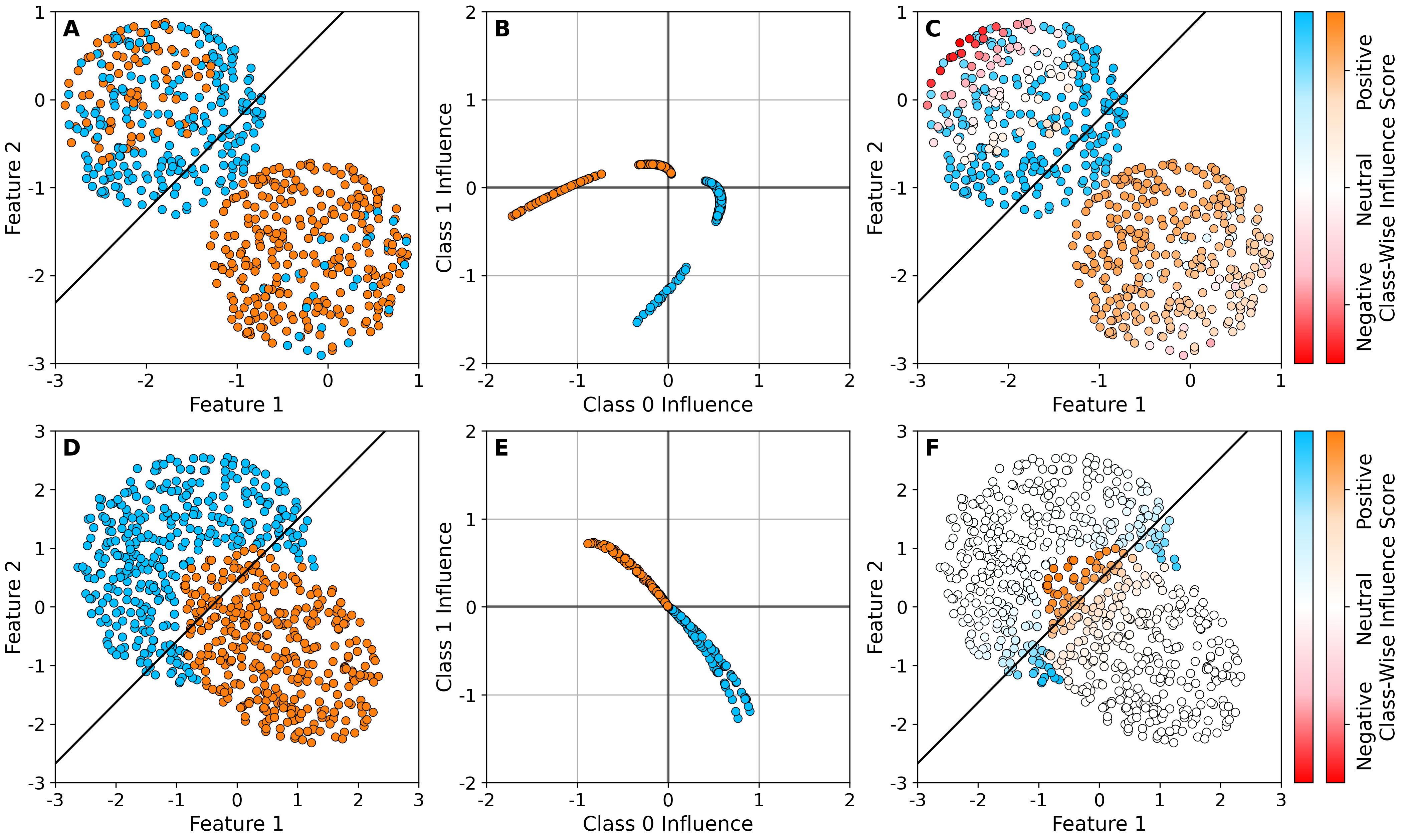}\vspace{-2mm}
    \caption{Validation of our category-wise influence function methods for analyzing the Pareto frontier on two synthetic binary classification datasets with logistic regression. Subfigures \textbf{A-C} showcase results on a synthetic dataset that is linearly separable and contains noisy detrimental training samples, where performance can improve by mislabeled sample removal. Subfigures \textbf{D-F} detail results for our method on a non-linearly separable dataset without any noisy samples, where performance improvements cannot be made for either class without sacrificing performance for the other. Subfigures \textbf{A} and \textbf{D} showcase the distribution of training samples for each of the two datasets with blue and orange denoting the ground-truth class labels. Subfigures \textbf{B} and \textbf{E} showcase the category-wise influence score distribution for both datasets. Further, subfigures \textbf{C} and \textbf{F} map the influence values to the training samples using color intensity in accordance with class colors to denote the influence magnitudes, where the original class color means positive and red means negative. }
    \label{fig:toy}\vspace{-4mm}
\end{figure*}

\section{Experimental Setup}
\label{sec:setup}

\looseness-1 Having described our proposed category-wise influence functions and the \textsc{Pareto-LP-GA} framework, we now outline the datasets and model configurations used to evaluate our approach. We conduct experiments on both synthetic and real-world settings to assess the validity of our performance-ceiling criterion and the effectiveness of our sample reweighting method. Specifically, we designed synthetic datasets to isolate and validate the performance-ceiling criterion under perfectly controlled conditions of noise and non-linear separability. For our real-world settings, we specifically selected benchmark datasets that naturally exhibit major class-wise tradeoffs and delayed convergence during training, ensuring there was sufficient room to demonstrate actual Pareto improvements. In the subsequent sections, we describe the datasets and models used in our experiments.

\subsubsection{Synthetic Datasets}
We evaluate our approach on two synthetic binary classification datasets. 
The first is a linearly separable dataset containing noisy detrimental 
training samples. It consists of 300 blue class samples and 300 orange 
class samples, generated using a circular uniform distribution. Noise was 
added to the training set by choosing random points from each group, 50 
from blue and 20 from orange, and then flipping their labels.

The second is a non-linearly separable dataset where performance 
improvements for both categories cannot be jointly made. It consists of 
350 samples for the blue category and 350 samples for the orange category, 
generated using a circular uniform distribution. For the orange samples, 
the radius of the distribution was changed depending on the angle from the 
center. This dataset contains no mislabeled samples.

\subsubsection{Real-world Datasets}
We use five widely adopted real-world benchmark datasets: three image datasets (\textit{CIFAR-10}~\citep{krizhevsky2009learning}, \textit{STL-10}~\citep{pmlr-v15-coates11a}), and a subset of ImageNet~\cite{deng2009imagenet} and two text datasets 
(\textit{Emotion}~\citep{saravia2018carer} and 
\textit{AG$\_$News}~\citep{zhang2015character}).  

We train \textit{bert-base-cased}~\citep{DBLP:journals/corr/abs-1810-04805} 
for the \textit{Emotion} and \textit{AG\_News} NLP datasets, and 
\textit{ResNet-9}~\citep{he2016deep} for the \textit{CIFAR10} and 
\textit{STL-10} vision datasets in our experiments. Additionally, we experiment with TinyViT~\citep{wu2022tinyvit} architecture to evaluate architectural generalization. 


\vspace{-2mm}
\section{Synthetic Data Verification}
\looseness-1 We now analyze the efficacy of our criterion on the model's performance ceiling with two synthetic binary classification datasets using logistic regression, as shown in Figure \ref{fig:toy}. The top row subfigures \textbf{A-C} denote a linearly separable dataset which consists of noisy samples that are mislabeled. Clearly, removing these noisy samples should improve performance and our category-wise influence functions should reflect their detrimental influence for both classes/categories. The large majority of non-noisy samples should positively influence one of the classes and negatively influence the other class. Hence, removing these samples should sacrifice performance for one of the classes and should be reflected in the category-wise influence distribution. As can be observed in subfigures \textbf{B} and \textbf{C} which showcase the category-wise influences and the samples corresponding to those influence values, respectively, this is indeed the case. Essentially, the mislabeled noisy samples from both categories possess negative influence for both categories, and hence, removing those should improve performance as expected. Additionally, a large number of non-noisy samples are positively/negatively influential for the blue/orange class (and vice-versa), validating our criterion further.

\looseness-1The bottom row consisting of subfigures \textbf{D-F} denotes a dataset that is non-linearly separable where performance improvements for both categories/classes cannot be jointly made. In the ideal scenario, the Pareto frontier obtained via our category-wise influence functions should indicate that samples are positively influential for one class and negatively influential for the other class. Moreover, the samples with the maximum influence magnitude are those that appear around the decision boundary and ones that will inadvertently be misclassified due to the linear logistic regression classifier. In subfigures \textbf{E} and \textbf{F}, we can see that this holds true with the Pareto frontier visualized. The influence vectors of all training samples form a roughly straight line. As we demonstrate through these results, category-wise influence vectors can reveal the Pareto frontier for the two datasets accurately, and help users/developers make tradeoffs as required depending on the needs of their given application. We also conduct the sample removal experiments on this synthetic dataset, which can be found in Appendix~\ref{app:syn}.


\vspace{-2mm}
\section{Real-world Data Experiments}
\looseness-1 Now, we present our experimental results on real-world datasets in two parts: a validation of the effectiveness of category-wise influence functions and an evaluation of our \textsc{Pareto-LP-GA} for Pareto performance improvement.

\subsection{Category-wise Influence Functions}
\looseness-1 We evaluate the category-wise influence using four widely adopted benchmark datasets with 4-10 categories: two vision datasets (\textit{CIFAR-10}~\citep{krizhevsky2009learning} and \textit{STL-10}~\citep{pmlr-v15-coates11a}) and two text datasets (\textit{Emotion}~\citep{saravia2018carer} and \textit{AG$\_$News}~\citep{zhang2015character}). For calculating category-wise sample influence vectors, we employ EKFAC~\citep{grosse2023studying} due to its fast implementation on deep models. 


\begin{figure*}[t]
    \centering
    \includegraphics[width=0.98\linewidth]{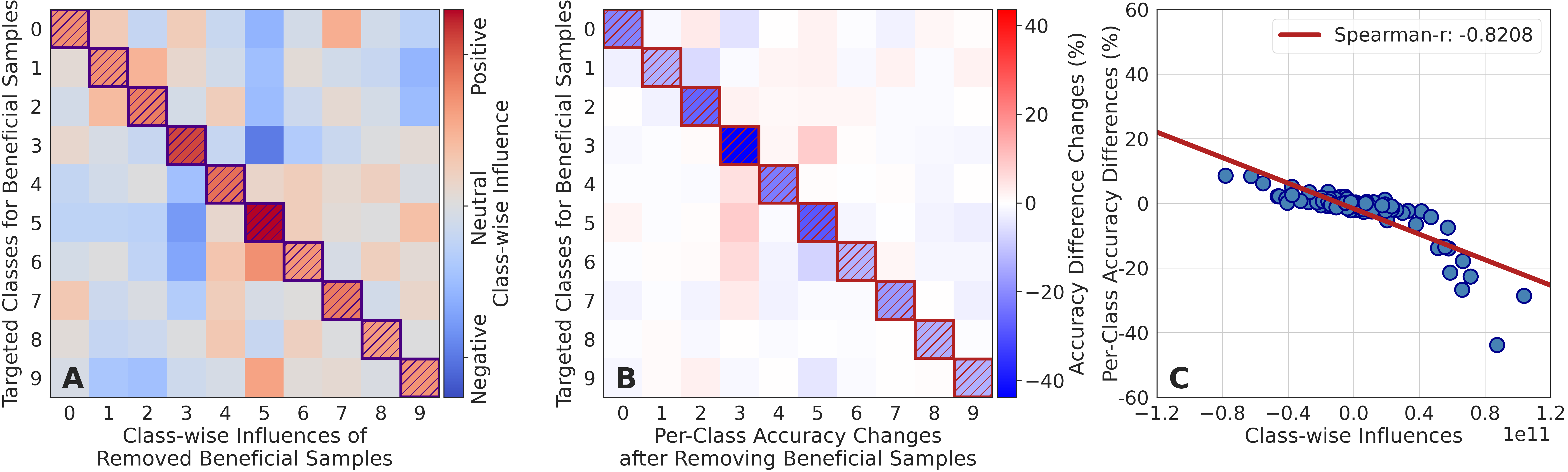}
    \includegraphics[width=0.98\linewidth]{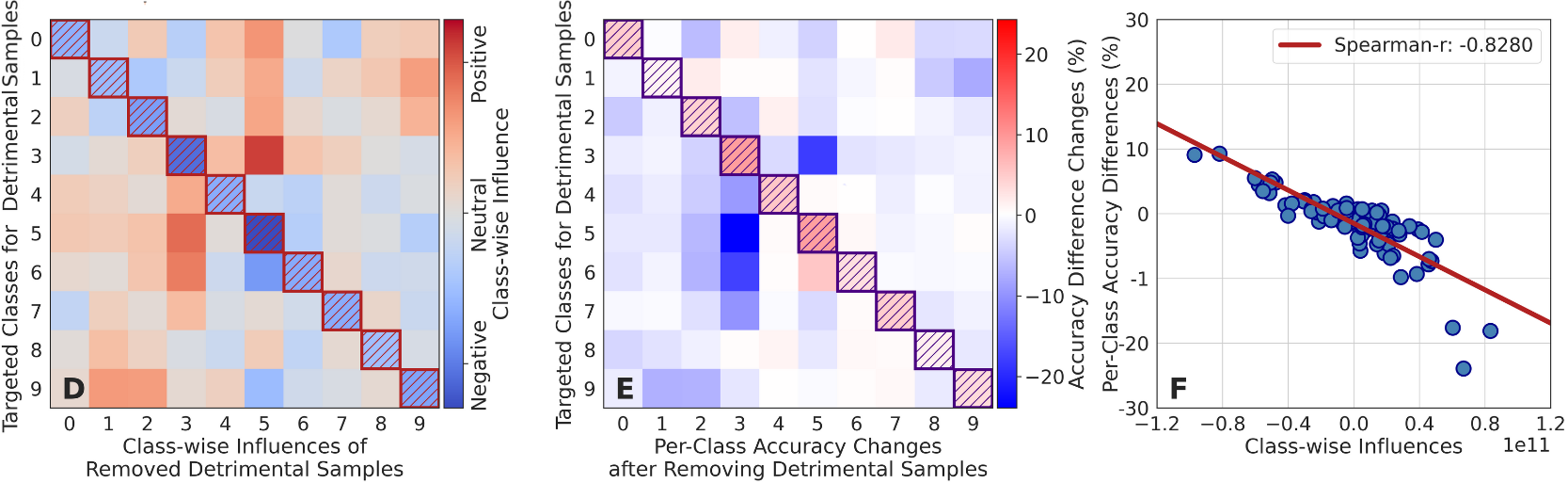}\vspace{-2mm}
    \caption{Category-wise influence on \textit{CIFAR10}~\citep{krizhevsky2009learning}. Subfigures \textbf{(A, D)} denote predicted category-wise influence and \textbf{(B, E)} actual accuracy changes with beneficial/detrimental sample removal. Subfigures \textbf{(C, F)} denote scatter plots between predicted influence and actual performance shifts.}
    \label{fig:cifar}\vspace{-2mm}
\end{figure*}

\begin{figure*}[htb!]
    \centering
    \includegraphics[width=0.98\linewidth]{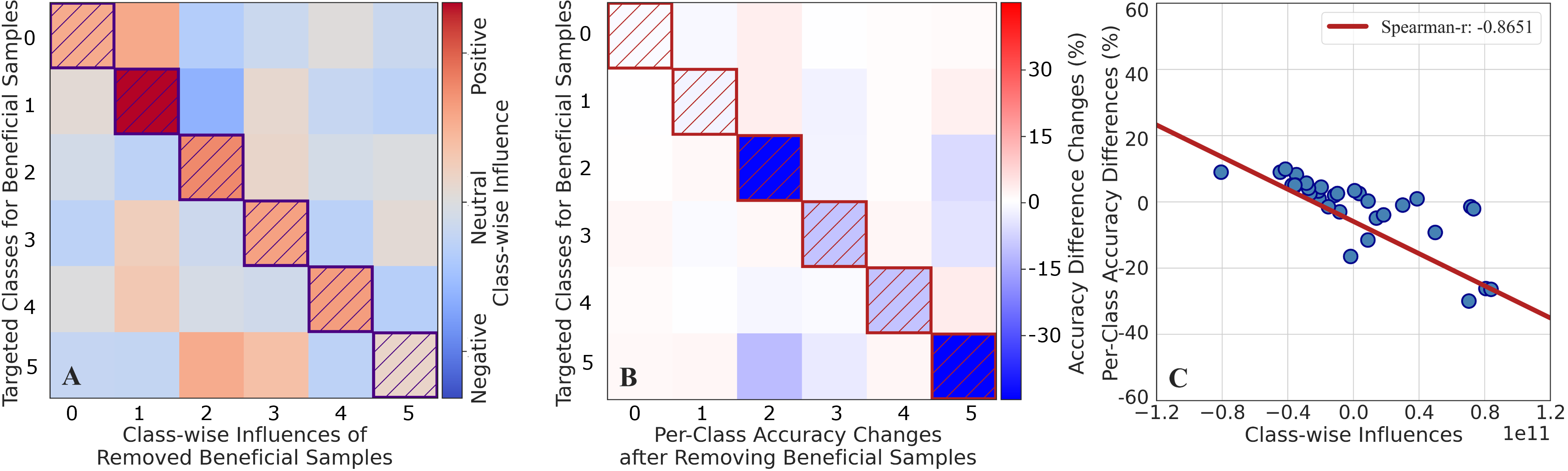}
    \includegraphics[width=0.98\linewidth]{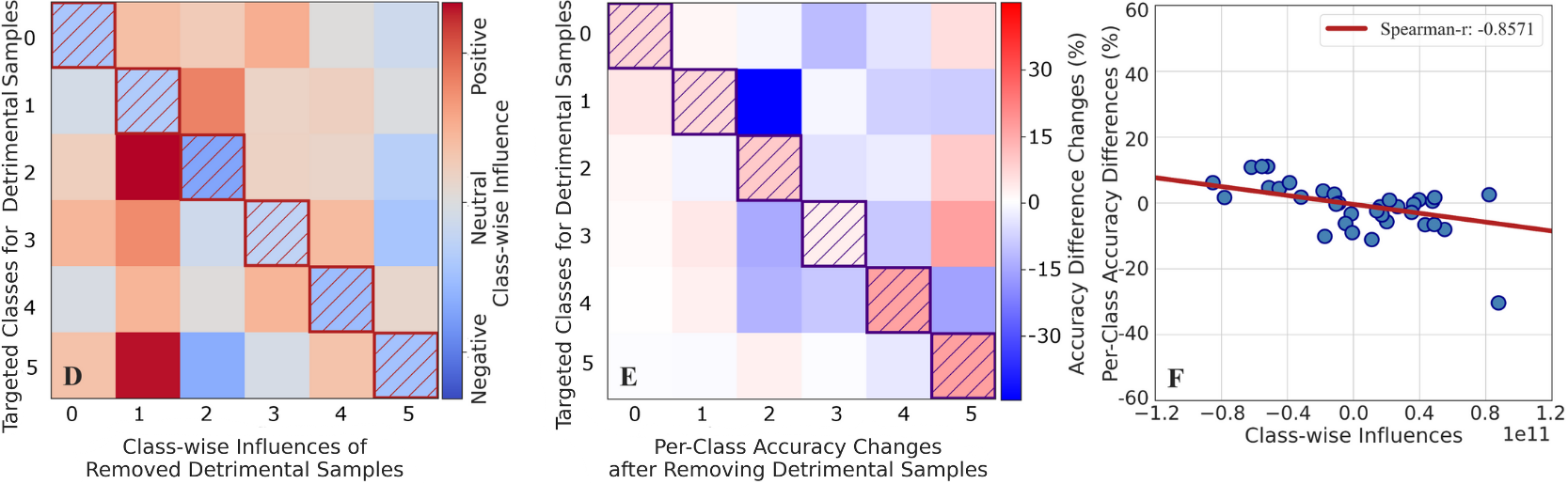}\vspace{-2mm}
    \caption{Category-wise influence on \textit{Emotion}~\citep{saravia2018carer} text dataset. This figure follows the same setup and caption format as Fig.~\ref{fig:cifar}.}\vspace{-5mm}
    \label{fig:emotion}
\end{figure*}

\looseness-1First, we evaluate whether category-wise influence serves as an effective indicator for measuring performance changes. To achieve this, we select the top 10\% of beneficial and detrimental samples for each category, remove these samples from the training set, and retrain the model to observe the performance change. Figures~\ref{fig:cifar} and~\ref{fig:emotion} illustrate the performance changes for \textit{CIFAR-10} and \textit{Emotion}. Similar phenomena are observed for \textit{STL-10} and \textit{AG$\_$News} (See Appendix~\ref{app:real}). In these figures, Subfigures \textbf{A} and \textbf{D} illustrate the cumulative influence values of the removed training samples for beneficial and detrimental samples, respectively. Subfigures \textbf{B} and \textbf{E} display the corresponding performance changes after retraining without these samples. The results are presented as heat maps, with diagonal blocks representing the targeted categories, highlighted using bold boundaries and distinct textures.

\begin{figure*}[htb!]
    \centering
    \includegraphics[width=0.95\linewidth]{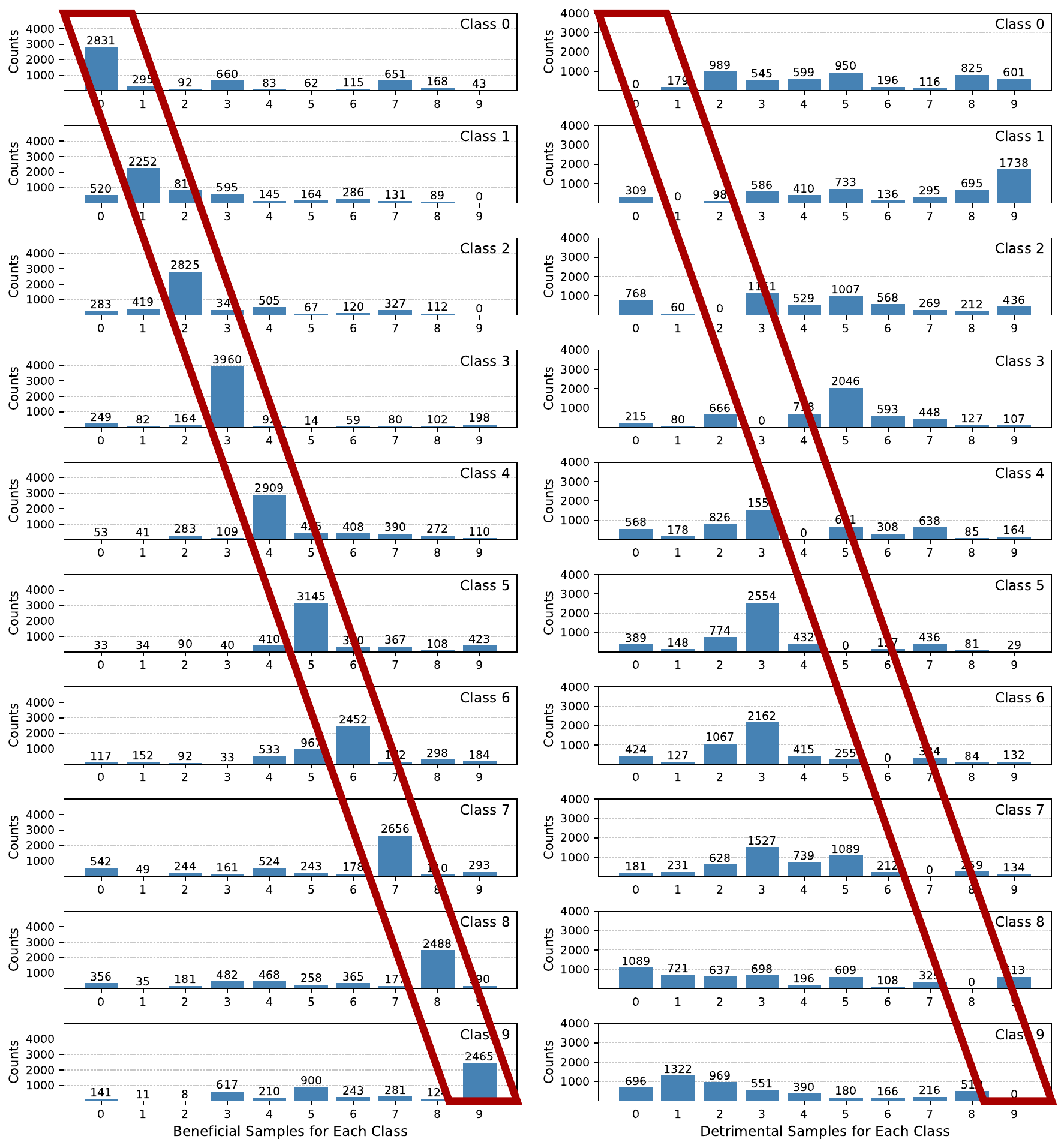}
    \caption{The numbers of beneficial and detrimental samples for each classes on \textit{CIFAR10}~\citep{krizhevsky2009learning} image dataset.}\vspace{-2mm}
    \label{fig:number_benef_detri}
\end{figure*}

The phenomena observed in the diagonal blocks reveal clear and consistent patterns: removing beneficial samples with positive influence on a target category leads to a performance drop in the corresponding category, while removing detrimental samples with negative influence results in a performance increase. This confirms the effectiveness of category-wise influence in predicting performance changes for both beneficial samples and detrimental samples. 
Beyond the diagonal blocks, the patterns are more mixed, as the selected samples based on one category may have varying impacts on other categories, which can be further verified by Fig.~\ref{fig:number_benef_detri}. To further analyze these non-diagonal patterns, we plot scatter diagrams of the cumulative influence for each category against their performance changes in Subfigures \textbf{C} and \textbf{F}. Subfigure \textbf{C} is derived from Subfigures \textbf{A} and \textbf{B}, while Subfigure \textbf{F} is based on Subfigures \textbf{D} and \textbf{E}. The Spearman correlation coefficient exceeding 0.8 indicates a strong relationship between sample influence and performance change, not only within the target category but also across categories. This demonstrates that category-wise influence functions can serve as an effective indicator for inferring performance changes. Moreover, they can be utilized to evaluate whether a classifier has reached its performance ceiling across different categories, providing valuable insights for targeted model improvement.

\vspace{-2mm}
\subsection{Performance Ceiling Check and Improvement}


\begin{table*}[h]
\centering
\caption{Comparison of category-wise accuracies for performance improvement in the \textit{Direct Improvement} (left) and \textit{Course Correction} (right) settings. Target classes are highlighted in blue. As can be observed, performance in target categories increases significantly while non-target classes see minimal reductions (or potential gains).}
\par\vspace{-1mm}
\label{tab:DI-CC-comparison}
\begin{minipage}[t]{0.44\textwidth}
\centering
\label{tab:DI-11}
\resizebox{0.95\textwidth}{!}{%
\begin{tabular}{c*{2}{S[table-format=1.3]}c}
\hline
\textbf{Category} & \textbf{Epoch-10} & \textbf{Epoch-11 (DI)} & \textbf{Change (\%)} \\ \hline
\rowcolor{blue!8}
0 & 0.699 & 0.811 & \textbf{\textcolor{ForestGreen}{+16.02}} \\ 
1 & 0.888 & 0.881 & -0.78 \\
\rowcolor{blue!8}
2 & 0.667 & 0.743 & \textbf{\textcolor{ForestGreen}{+11.39}} \\
3 & 0.647 & 0.632 & -2.31 \\
4 & 0.729 & 0.720 & -1.2 \\ 
5 & 0.755 & 0.755 & \textcolor{ForestGreen}{+0.00} \\
6 & 0.802 & 0.845 & \textcolor{ForestGreen}{+5.73} \\
7 & 0.849 & 0.848 & -0.11 \\ 
8 & 0.948 & 0.920 & -2.90 \\
9 & 0.817 & 0.818 & \textcolor{ForestGreen}{+0.12} \\ \hline
\end{tabular}}

\end{minipage}
\hfill
\begin{minipage}[t]{0.54\textwidth}
\centering
\label{tab:CC-16}
\resizebox{0.95\textwidth}{!}{%
\begin{tabular}{c*{2}{S[table-format=1.3]}cc}
\hline
\textbf{Category} & \textbf{Epoch-15} & \textbf{Epoch-16} & \textbf{Epoch-16 (CC)} & \textbf{Change (\%)} \\ \hline
0 & 0.876 & 0.876 & 0.889 & \textcolor{ForestGreen}{+1.48} \\ 
1 & 0.866 & 0.868 & 0.870 & \textcolor{ForestGreen}{+0.23} \\
2 & 0.651 & 0.741 & 0.736 & -0.67 \\
3 & 0.582 & 0.678 & 0.677 & -0.14 \\
4 & 0.729 & 0.783 & 0.785 & \textcolor{ForestGreen}{+0.25} \\ 
\rowcolor{blue!8}
5 & 0.821 & 0.785 & 0.798 & \textbf{\textcolor{ForestGreen}{+1.65}} \\
6 & 0.859 & 0.859 & 0.855 & -0.46 \\
\rowcolor{blue!8}
7 & 0.885 & 0.837 & 0.848 & \textbf{\textcolor{ForestGreen}{+1.31}} \\ 
8 & 0.909 & 0.929 & 0.917 & -1.29 \\
\rowcolor{blue!8}
9 & 0.917 &  0.864 & 0.888 & \textbf{\textcolor{ForestGreen}{+2.77}} \\ \hline
\end{tabular}}
\end{minipage}\vspace{-3mm}
\end{table*}



We evaluate our \textsc{Pareto-LP-GA} method using the \textit{CIFAR10} dataset and ResNet model to showcase performance improvements for both \textit{Direct Improvement} (\textit{DI})  and \textit{Course Correction} (\textit{CC}) settings. The reason we choose \textit{CIFAR10} for experiments is that during training multiple epochs exhibit major Pareto tradeoffs across categories, while ensuring the room for potential improvement. In contrast, for our text datasets (\textit{Emotion} and \textit{AG\_News}), the NLP models achieved accuracies exceeding 90\% across all classes within the first epoch, leading to little room for Pareto frontier improvement. Similarly, \textit{STL-10} consists of \textit{cleaner} images than \textit{CIFAR10}, generally leading to better performance.

\looseness-1 Before demonstrating performance improvements, we first examine whether the influence vectors of the training samples approximately lie on a hyperplane. We apply Principal Component Analysis \citep{wold1987principal} and compute the explained variance ratio of the first principal component. Across all targeted cases, this ratio consistently exceeds 0.2, indicating that the influence vectors do not fit a hyperplane and suggesting room for Pareto improvement.

\begin{table*}[htb!]
\centering
\caption{Comparison of category-wise accuracies for \textbf{CIFAR-10 Final Epoch (Epoch-19)} improvement in \textit{Direct Improvement} (left) and \textit{Course Correction} (right). Target classes are highlighted in blue.}
\par\vspace{2mm}
\label{tab:cifar-final-comparison}
\begin{minipage}[t]{0.44\textwidth}
\centering
\resizebox{0.9\textwidth}{!}{%
\begin{tabular}{c*{2}{S[table-format=1.3]}c}
\hline
\textbf{Class} & \textbf{Epoch-18} & \textbf{Epoch-19 (DI)} & \textbf{Change (\%)} \\ \hline
0 & 0.904 & 0.895 & -1.00 \\ 
1 & 0.838 & 0.868 & \textcolor{ForestGreen}{+3.50} \\
\rowcolor{blue!8}
2 & 0.703 & 0.704 & \textbf{\textcolor{ForestGreen}{+0.14}} \\
\rowcolor{blue!8}
3 & 0.724 & 0.733 & \textbf{\textcolor{ForestGreen}{+1.24}} \\
\rowcolor{blue!8}
4 & 0.781 & 0.810 & \textbf{\textcolor{ForestGreen}{+3.71}} \\ 
\rowcolor{blue!8}
5 & 0.758 & 0.795 & \textbf{\textcolor{ForestGreen}{+4.88}} \\
6 & 0.869 & 0.879 & \textcolor{ForestGreen}{+1.15} \\
7 & 0.832 & 0.819 & -1.56 \\ 
8 & 0.918 & 0.919 & \textcolor{ForestGreen}{+0.11} \\
9 & 0.881 & 0.898 & \textcolor{ForestGreen}{+1.92} \\ \hline
\end{tabular}}
\end{minipage}
\hfill
\begin{minipage}[t]{0.54\textwidth}
\centering
\resizebox{0.9\textwidth}{!}{%
\begin{tabular}{c*{3}{S[table-format=1.3]}c}
\hline
\textbf{Class} & \textbf{Epoch-18} & \textbf{Epoch-19} & \textbf{Epoch-19 (CC)} & \textbf{Change (\%)} \\ \hline
0 & 0.904 & 0.897 & 0.897 & +0.00 \\ 
1 & 0.838 & 0.871 & 0.863 & -0.92 \\
2 & 0.703 & 0.725 & 0.711 & -1.93 \\
\rowcolor{blue!8}
3 & 0.724 & 0.700 & 0.703 & \textbf{\textcolor{ForestGreen}{+0.42}} \\
4 & 0.781 & 0.786 & 0.784 & -0.25 \\ 
5 & 0.758 & 0.810 & 0.814 & \textcolor{ForestGreen}{+0.49} \\
6 & 0.869 & 0.872 & 0.879 & \textcolor{ForestGreen}{+0.80} \\
\rowcolor{blue!8}
7 & 0.832 & 0.827 & 0.831 & \textbf{\textcolor{ForestGreen}{+0.48}} \\ 
8 & 0.918 & 0.923 & 0.920 & -0.32 \\
9 & 0.881 & 0.892 & 0.905 & \textcolor{ForestGreen}{+1.56} \\ \hline
\end{tabular}}
\end{minipage}\vspace{-2mm}
\end{table*}

\begin{table*}[htb!]
\centering
\caption{Comparison of category-wise accuracies for \textbf{CIFAR-10 Course Correction (Epoch-15)} between our method and a \textit{target-class reweighting baseline} (ratio of average accuracy to target class accuracy). Target classes (C-5, C-6) are highlighted in blue.}
\par\vspace{2mm}
\label{tab:cifar-reweight-comparison}
\resizebox{0.9\textwidth}{!}{%
\begin{tabular}{c*{3}{S[table-format=1.3]}c S[table-format=1.3] c}
\hline
\textbf{Class} & \textbf{Epoch-14} & \textbf{Epoch-15} & \textbf{Epoch-15 (CC)} & \textbf{Change (CC) (\%)} & \textbf{Epoch-15 (Reweight)} & \textbf{Change (Reweight) (\%)} \\ \hline
C-0 & 0.819 & 0.826 & 0.826 & \textcolor{ForestGreen}{+0.00}                          & 0.820 & -0.726 \\
C-1 & 0.899 & 0.906 & 0.900 & -0.66                          & 0.906 & \textcolor{ForestGreen}{+0.00}  \\
C-2 & 0.709 & 0.750 & 0.765 & \textcolor{ForestGreen}{+2.00} & 0.746 & -0.533 \\
C-3 & 0.635 & 0.673 & 0.690 & \textcolor{ForestGreen}{+2.50} & 0.679 & \textcolor{ForestGreen}{+0.891} \\
C-4 & 0.703 & 0.724 & 0.717 & -0.97                          & 0.731 & \textcolor{ForestGreen}{+0.967} \\
\rowcolor{blue!8}
C-5 & 0.790 & 0.768 & 0.780 & \textbf{\textcolor{ForestGreen}{+1.56}} & 0.770 & \textbf{\textcolor{ForestGreen}{+0.26}} \\
\rowcolor{blue!8}
C-6 & 0.848 & 0.823 & 0.827 & \textbf{\textcolor{ForestGreen}{+0.55}} & 0.820 & -0.365 \\
C-7 & 0.862 & 0.879 & 0.858 & -2.38                          & 0.883 & \textcolor{ForestGreen}{+0.455} \\
C-8 & 0.948 & 0.932 & 0.926 & -0.64                          & 0.941 & \textcolor{ForestGreen}{+0.965} \\
C-9 & 0.841 & 0.890 & 0.887 & -0.33                          & 0.989 & -0.112 \\ \hline
\end{tabular}}
\vspace{-2mm}
\end{table*}

\looseness-1We present results for \textit{DI} and \textit{CC} in Table~\ref{tab:DI-CC-comparison}. 
For \textit{DI} in \textit{CIFAR10}, we identify two categories (0 and 2) with relatively lower accuracy after observing performance at \textit{Epoch 10} (0.699 for class-0, and 0.667 for class-2). These two classes will constitute our target categories. Then, we employ \textsc{Pareto-LP-GA} to obtain a weight set and apply weighted training to achieve \textit{Epoch 11}. The results demonstrate that \textsc{Pareto-LP-GA} significantly enhances performance on the target categories, achieving improvements of 16.02\% and 11.39\% for classes 0 and 2, respectively. Notably, several non-target categories also experienced performance gains, such as category 6, which improved by 5.73\%. However, for the non-target categories, performance degradation remains very minimal, showcasing the performance ceiling of the classifier via the Pareto improvement.

\looseness-1For \textit{CC}, we identify \textit{Epoch 16} as detrimental for classes 5, 7, and 9, where accuracies decline substantially (from \textit{Epoch 15}). That is, between Epochs 15 and 16, performance for class 5 drops from 0.821 $\rightarrow$ 0.785; for class 7 drops from 0.885 $\rightarrow$ 0.837; and for class 9, drops from 0.917 $\rightarrow$ 0.864. Thus, these classes constitute our target categories for \textit{CC}. We obtain the weight set using \textsc{Pareto-LP-GA} and after applying weighted training, we seek to obtain a new Epoch 16 that optimizes for the Pareto-optimal class tradeoffs as desired. This can be observed in our results -- accuracy improves by 1.65\%, 1.31\%, and 2.77\% in target categories 5, 7, and 9, respectively, reversing the original trends and bringing slight improvements. Additionally, class-0, class-1, and class-4 experienced 1.48\%, 0.23\%, and 0.25\% performance enhancement, respectively, and there were only modest accuracy reductions in other categories ($\leq 1.29\%$). Overall, our proposed \textsc{Pareto-LP-GA} method demonstrates its ability to enhance training performance in specific target categories with minor accuracy tradeoffs in non-target categories.

\looseness-1 We further apply our method at the final training stage (Epoch 19) of CIFAR-10 to demonstrate its effectiveness as a "final polish" for deployed models. Table \ref{tab:cifar-final-comparison} demonstrates the results. In the \textit{Direct Improvement} setting, we targeted classes with lower relative performance (Classes 2, 3, 4, 5). We achieved significant gains, such as a +4.88\% increase for Class 5 and +3.71\% for Class 4. In the \textit{Course Correction} setting, we targeted Classes 3 and 7, which experienced minor regression in the standard final epoch; our method successfully recovered their performance without retraining the model from scratch.

\looseness-1 To verify whether the observed improvements can be achieved through a simple optimization heuristic, we further compare our method against a target-class reweighting baseline, which rescales the loss contribution of target classes by the ratio of average accuracy to target class accuracy. Table~\ref{tab:cifar-reweight-comparison} reports results under the \textit{CC} setting on CIFAR-10 (Epoch-15) with target classes C-5 and C-6. While the reweighting baseline yields \textit{inconsistent gains} and in several cases harms both target and non-target classes (e.g., -0.726\% for C-0, -0.533\% for C-2, -0.365\% for C-6), our method achieves more \textit{stable and balanced improvements}, successfully recovering the target classes without degrading overall model performance.

\begin{table}[t]
\centering
\caption{Average accuracy across all classes for \textit{multi-epoch Dynamic DI} on CIFAR-10 (ResNet-9), comparing standard training against \textsc{Pareto-LP-GA} applied over two consecutive epochs.}
\par\vspace{1mm}
\label{tab:cifar-dynamic-di}
\resizebox{0.49\textwidth}{!}{%
\begin{tabular}{lcc}
\hline
\textbf{Epoch} & \textbf{Without \textsc{Pareto-LP-GA}} & \textbf{With \textsc{Pareto-LP-GA} (DI)} \\ \hline
15 & $0.672 \pm 0.191$ & $0.672 \pm 0.191$ \\
16 & $0.655 \pm 0.154$ & \textcolor{ForestGreen}{$\mathbf{0.747 \pm 0.085}$} \\
17 & $0.683 \pm 0.148$ & \textcolor{ForestGreen}{$\mathbf{0.751 \pm 0.083}$} \\ \hline
\end{tabular}}
\vspace{-6mm}
\end{table}

\looseness-1 To evaluate whether our framework remains effective when target classes are updated per epoch based on current class-wise performance, we extend it to a \textit{dynamic setting} with adaptively selected targets. We conduct an additional multi-epoch \textit{Direct Improvement} experiment in which \textsc{Pareto-LP-GA} is applied over two consecutive epochs (Epoch 16 and Epoch 17) with adaptively chosen targets. In Epoch 16 (targets: C-3, C-4, C-5), per-class accuracy improves from 0.564, 0.595, 0.195 to 0.636, 0.765, 0.589, respectively. In Epoch 17 (targets: C-3, C-5), accuracy further improves from 0.636, 0.589 to 0.658, 0.728.

\looseness-1 Table~\ref{tab:cifar-dynamic-di} summarizes the average accuracy across all classes. With \textsc{Pareto-LP-GA}, average accuracy increases from 0.672 to 0.747 (Epoch 15~$\to$~16) and continues to improve to 0.751 by Epoch 17, while the standard deviation decreases from 0.191 to 0.083, indicating significantly more \textit{balanced performance across classes}. In contrast, the standard training without \textsc{Pareto-LP-GA} shows \textit{no consistent improvement} and maintains \textit{higher variance} across epochs. Overall, these results demonstrate that applying \textsc{Pareto-LP-GA} across consecutive epochs leads to \textit{gradual improvement} with better \textit{class-wise balance}. We have presented the results on the CIFAR-10 dataset using the ResNet-9 model. However, we also conduct experiments on a more complex dataset and a different model architecture to evaluate the generalization capability of \textsc{Pareto-LP-GA}. We now discuss the performance of \textsc{Pareto-LP-GA} on the new dataset and model architecture.

\subsection{Generalization Check}

To evaluate our method on larger and more complex data, we conducte experiments on a subset of ImageNet (55k training samples, 11 Superclasses \citep{tsipras2020imagenetimageclassificationcontextualizing}). We illustrate two scenarios in Table \ref{tab:imagenet-comparison}: \textit{Direct Improvement} (Left) targeting Superclasses 6, 8, 9, and 10; and \textit{Course Correction} (Right) targeting Superclasses 0, 2, 3, and 8 to recover performance lost during standard training.

\begin{table*}[t]
\centering
\caption{Comparison of category-wise accuracies for the \textbf{ImageNet Subset} in \textit{Direct Improvement} (left) and \textit{Course Correction} (right). Target classes are highlighted in blue.}
\par\vspace{2mm}
\label{tab:imagenet-comparison}

\resizebox{0.9\textwidth}{!}{%
\begin{minipage}[t]{0.44\textwidth}
\centering
\resizebox{0.9\textwidth}{!}{%
\begin{tabular}{c*{2}{S[table-format=2.1]}c}
\hline
\textbf{Class} & \textbf{Epoch-18} & \textbf{Epoch-19 (DI)} & \textbf{Change} \\ \hline
0 & 49.7 & 57.0 & \textcolor{ForestGreen}{+7.3} \\ 
1 & 43.0 & 37.9 & -5.1 \\
2 & 54.4 & 49.7 & -4.7 \\
3 & 52.1 & 53.7 & \textcolor{ForestGreen}{+1.6} \\
4 & 54.6 & 48.1 & -6.5 \\ 
5 & 41.0 & 40.9 & -0.1 \\
\rowcolor{blue!8}
6 & 12.9 & 18.6 & \textbf{\textcolor{ForestGreen}{+5.7}} \\
7 & 43.1 & 51.5 & \textcolor{ForestGreen}{+8.4} \\ 
\rowcolor{blue!8}
8 & 31.0 & 32.8 & \textbf{\textcolor{ForestGreen}{+1.8}} \\
\rowcolor{blue!8}
9 & 32.1 & 32.5 & \textbf{\textcolor{ForestGreen}{+0.4}} \\ 
\rowcolor{blue!8}
10 & 26.7 & 40.4 & \textbf{\textcolor{ForestGreen}{+13.7}} \\ \hline
\end{tabular}}
\end{minipage}
\hfill
\begin{minipage}[t]{0.54\textwidth}
\centering
\resizebox{0.93\textwidth}{!}{%
\begin{tabular}{c*{3}{S[table-format=2.1]}c}
\hline
\textbf{Class} & \textbf{Epoch-18} & \textbf{Epoch-19} & \textbf{Epoch-19 (CC)} & \textbf{Change} \\ \hline
\rowcolor{blue!8}
0 & 49.7 & 43.6 & 56.7 & \textbf{\textcolor{ForestGreen}{+13.1}} \\ 
1 & 43.0 & 41.9 & 34.3 & -7.6 \\
\rowcolor{blue!8}
2 & 54.4 & 41.5 & 49.6 & \textbf{\textcolor{ForestGreen}{+8.1}} \\
\rowcolor{blue!8}
3 & 52.1 & 43.0 & 51.1 & \textbf{\textcolor{ForestGreen}{+8.1}} \\
4 & 54.6 & 61.4 & 51.6 & -9.8 \\ 
5 & 41.0 & 43.6 & 37.3 & -6.3 \\
6 & 12.9 & 17.0 & 17.0 & +0.0 \\
7 & 43.1 & 54.8 & 57.0 & \textcolor{ForestGreen}{+2.2} \\ 
\rowcolor{blue!8}
8 & 31.0 & 26.4 & 30.6 & \textbf{\textcolor{ForestGreen}{+4.2}} \\
9 & 32.1 & 31.1 & 32.1 & \textcolor{ForestGreen}{+1.0} \\ 
10 & 26.7 & 32.7 & 42.7 & \textcolor{ForestGreen}{+10.0} \\ \hline
\end{tabular}}
\end{minipage}}\vspace{-4mm}
\end{table*}

\begin{figure*}[t]
    \centering
    \includegraphics[width=0.95\linewidth]{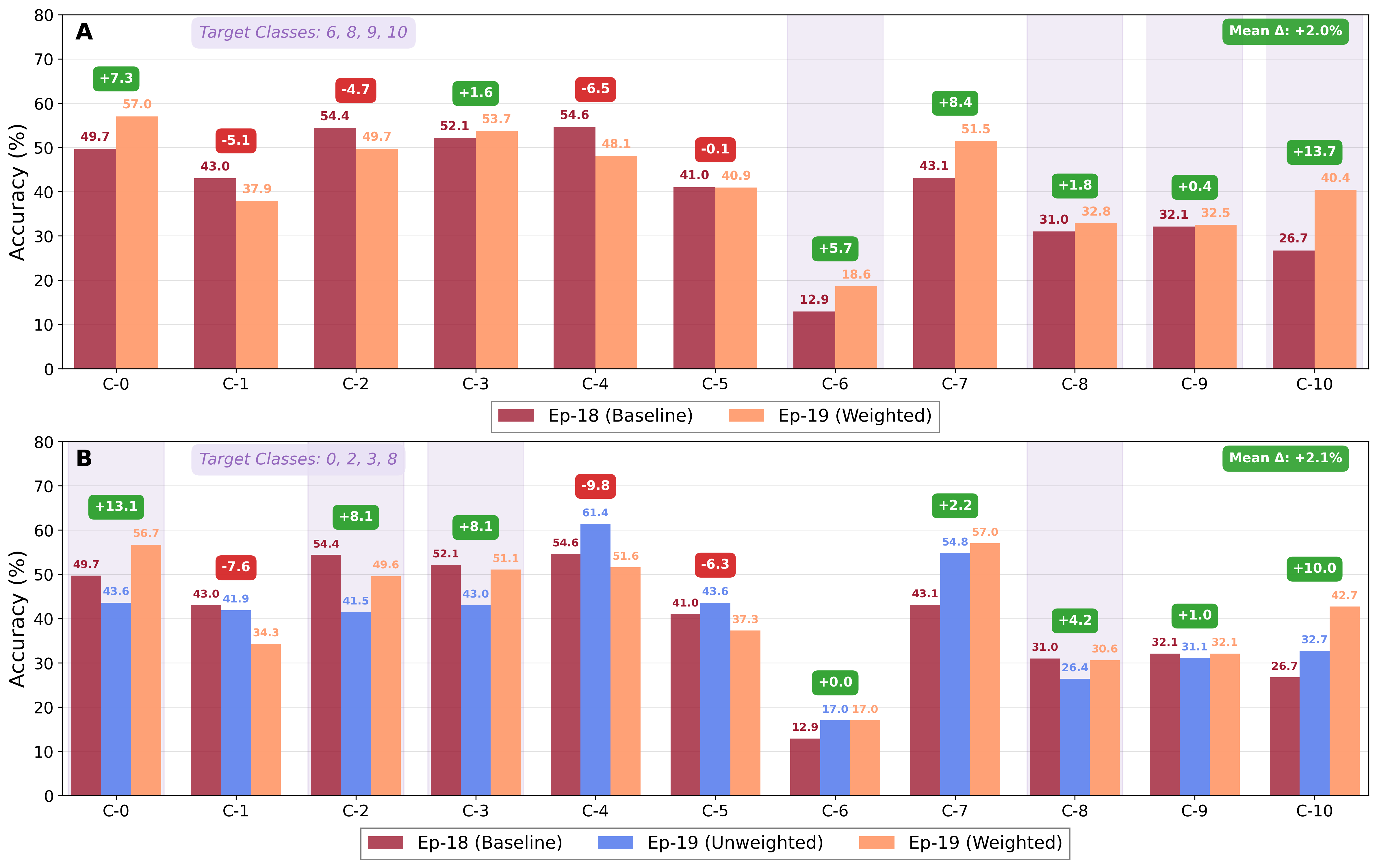}\vspace{-2mm}
    \caption{ImageNet performance shifts. \textbf{A} is the setting of DI and \textbf{B} is the setting of CC.}\vspace{2mm}
    \label{fig:imagenet}
\end{figure*}

In the Direct Improvement setting, we observe a gain for Class 10 (+13.7) and Class 7 (+8.4), boosting the overall average accuracy by 2.0. In the Course Correction setting, our method successfully reversed the performance degradation observed in standard Epoch 19, improving Class 0 by 13.1 relative to the baseline. We demonstrate the performance gains in a bar plot in Figure \ref{fig:imagenet}.

\begin{table}[t]
\centering
\caption{Category-wise accuracies for \textbf{CIFAR-10 Direct Improvement (Epoch-50)} using \textit{TinyViT} architecture. Target classes (C-2, C-3) are highlighted in blue.}
\par\vspace{2mm}
\label{tab:cifar-tinyvit-di}
\resizebox{0.38\textwidth}{!}{%
\begin{tabular}{c*{2}{S[table-format=1.3]}c}
\hline
\textbf{Class} & \textbf{Epoch-49} & \textbf{Epoch-50} & \textbf{Deviation (\%)} \\ \hline
C-0 & 0.605 & 0.604 & -0.16 \\
C-1 & 0.729 & 0.730 & \textcolor{ForestGreen}{+0.14} \\
\rowcolor{blue!8}
C-2 & 0.412 & 0.436 & \textbf{\textcolor{ForestGreen}{+5.80}} \\
\rowcolor{blue!8}
C-3 & 0.328 & 0.344 & \textbf{\textcolor{ForestGreen}{+4.80}} \\
C-4 & 0.477 & 0.474 & -0.63 \\
C-5 & 0.500 & 0.500 & \textcolor{ForestGreen}{+0.00} \\
C-6 & 0.691 & 0.679 & -1.70 \\
C-7 & 0.654 & 0.652 & -0.31 \\
C-8 & 0.728 & 0.724 & -0.55 \\
C-9 & 0.658 & 0.661 & \textcolor{ForestGreen}{+0.45} \\ \hline
\end{tabular}}
\vspace{-4mm}
\end{table}

\looseness-1 To assess the generalizability of our method beyond ResNet-based architecture, we replace ResNet-9 with TinyViT~\cite{wu2022tinyvit} and apply \textit{Direct Improvement} on CIFAR-10 over 50 epochs, targeting Class 2 and Class 3. Table~\ref{tab:cifar-tinyvit-di} reports the per-class accuracies at the final epoch (Epoch-50). Our method achieves substantial improvements on both target classes (+5.80\% for Class 2 and +4.80\% for Class 3), while non-target classes exhibit only \textit{minimal degradation}, confirming that the method transfers effectively across architectures.




\vspace{-2mm}
\section{Conclusion}
In this paper, we extended the conventional influence function to a category-wise influence function and introduced the concept of an influence vector, which quantifies the impact of each training sample across all categories. Building on this formulation, we analyzed whether a classifier has already reached its maximum potential performance: formally, the Pareto frontier across categories, and designed a linear programming–based sample reweighting framework for Pareto improvements. We validated the correctness of our performance-ceiling criterion through experiments on synthetic datasets, and further demonstrated the effectiveness of the proposed category-wise influence function on real-world datasets. Finally, we presented case studies that showcase how our sample reweighting approach can lead to tangible Pareto improvements across multiple categories.

\section*{Acknowledgments}
We gratefully acknowledge the support of the Google TPU Builders Program, which provided support and access to computational resources, and Google Tunix library for post-training functionality and flexibility to enable this work.

\bibliography{references}

@inproceedings{koh2017understanding,
  title={{Understanding black-box predictions via influence functions}},
  author={Koh, Pang Wei and Liang, Percy},
  booktitle={International Conference on Machine Learning},
  pages={1885--1894},
  year={2017}
}

@article{martin1986influence,
  title={{Influence Functionals for Time Series}},
  author={Martin, R Douglas and Yohai, Victor J},
  journal={The Annals of Statistics},
  year={1986},
  publisher={JSTOR}
}

@article{bae2024training,
  title={{Training Data Attribution via Approximate Unrolled Differentation}},
  author={Bae, Juhan and Lin, Wu and Lorraine, Jonathan and Grosse, Roger},
  journal={arXiv preprint arXiv:2405.12186},
  year={2024}
}

@article{tsipras2020imagenetimageclassificationcontextualizing,
   title={{From ImageNet to Image Classification: Contextualizing Progress on Benchmarks}}, 
      author={Dimitris Tsipras and Shibani Santurkar and Logan Engstrom and Andrew Ilyas and Aleksander Madry},
  journal={arXiv preprint arXiv:2005.11295},
  year={2020}
}

@book{cook1982residuals,
  title={{Residuals and Influence in Regression}},
  author={Cook, R Dennis and Weisberg, Sanford},
  year={1982},
  publisher={New York: Chapman and Hall}
}

@article{hampel1974influence,
  title={{The Influence Curve and Its Role in Robust Estimation}},
  author={Hampel, Frank R},
  journal={Journal of the American Statistical Association},
  year={1974},
  publisher={Taylor \& Francis}
}

@article{cohn1996active,
  title={{Active Learning with Statistical Models}},
  author={Cohn, David A and Ghahramani, Zoubin and Jordan, Michael I},
  journal={Journal of Artificial Intelligence Research},
  volume={},
  pages={},
  year={1996}
}

@inproceedings{chhabra2025outlier,
  title={{Outlier Gradient Analysis: Efficiently Identifying Detrimental Training Samples for Deep Learning Models}},
  author={Chhabra, Anshuman and Li, Bo and Chen, Jian and Mohapatra, Prasant and Liu, Hongfu},
  booktitle={International Conference on Machine Learning},
  year={2025}
}

@article{wold1987principal,
  title={{Principal Component Analysis}},
  author={Wold, Svante and Esbensen, Kim and Geladi, Paul},
  journal={Chemometrics and Intelligent Laboratory Systems},
  year={1987}
}

@article{dantzig2002linear,
  title={{Linear Programming}},
  author={Dantzig, George B},
  journal={Operations research},
  year={2002}
}

@inproceedings{han2020explaining,
  title={{Explaining Black Box Predictions and Unveiling Data Artifacts through Influence Functions}},
  author={Han, Xiaochuang and Wallace, Byron C and Tsvetkov, Yulia},
  booktitle={Annual Meeting of the Association for Computational Linguistics},
  pages={},
  year={2020}
}

@article{nguyen2022measure,
  title={{How to Measure Uncertainty in Uncertainty Sampling for Active Learning}},
  author={Nguyen, Vu-Linh and Shaker, Mohammad Hossein and H{\"u}llermeier, Eyke},
  journal={Machine Learning},
  volume={},
  number={},
  pages={},
  year={2022},
  publisher={Springer}
}

@article{paul2021deep,
  title={{Deep Learning on a Data Diet: Finding Important Examples Early in Training}},
  author={Paul, Mansheej and Ganguli, Surya and Dziugaite, Gintare Karolina},
  journal={Advances in Neural Information Processing Systems},
  volume={},
  pages={},
  year={2021}
}

@inproceedings{wei2015submodularity,
  title={{Submodularity in Data Subset Selection and Active Learning}},
  author={Wei, Kai and Iyer, Rishabh and Bilmes, Jeff},
  booktitle={International Conference on Machine Learning},
  pages={},
  year={2015},
  organization={}
}

@inproceedings{chhabra2022fair,
  title={{Fair Clustering Using Antidote Data}},
  author={Chhabra, Anshuman and Singla, Adish and Mohapatra, Prasant},
  booktitle={Algorithmic Fairness through the Lens of Causality and Robustness Workshop},
  pages={},
  year={2022},
  organization={}
}

@article{ilyas2022datamodels,
  title={{Datamodels: Predicting Predictions from Training Data}},
  author={Ilyas, Andrew and Park, Sung Min and Engstrom, Logan and Leclerc, Guillaume and Madry, Aleksander},
  journal={arXiv preprint arXiv:2202.00622},
  year={2022}
}

@inproceedings{yang2022dataset,
  title={{Dataset Pruning: Reducing Training Data by Examining Generalization Influence}},
  author={Yang, Shuo and Xie, Zeke and Peng, Hanyu and Xu, Min and Sun, Mingming and Li, Ping},
  booktitle={International Conference on Learning Representations},
  year={2022}
}

@inproceedings{schioppa2022scaling,
  title={{Scaling up Influence Functions}},
  author={Schioppa, Andrea and Zablotskaia, Polina and Vilar, David and Sokolov, Artem},
  booktitle={AAAI Conference on Artificial Intelligence},
  year={2022}
}

@article{pruthi2020estimating,
  title={{Estimating Training Data Influence by Tracing Gradient Descent}},
  author={Pruthi, Garima and Liu, Frederick and Kale, Satyen and Sundararajan, Mukund},
  journal={Advances in Neural Information Processing Systems},
  year={2020}
}

@article{yeh2018representer,
  title={{Representer Point Selection for Explaining Deep Neural Networks}},
  author={Yeh, Chih-Kuan and Kim, Joon and Yen, Ian En-Hsu and Ravikumar, Pradeep K},
  journal={Advances in Neural Information Processing Systems},
  year={2018}
}

@inproceedings{basu2020influence,
  title={{Influence {F}unctions in {D}eep {L}earning {A}re {F}ragile}},
  author={Basu, Samyadeep and Pope, Phil and Feizi, Soheil},
  booktitle={International Conference on Learning Representations},
  year={2020}
}

@inproceedings{chhabra2024what,
  title={{``What Data Benefits My Classifier?" Enhancing Model Performance and Interpretability through Influence-Based Data Selection}},
  author={Chhabra, Anshuman and Li, Peizhao and Mohapatra, Prasant and Liu, Hongfu},
  booktitle={International Conference on Learning Representations},
  year={2024}
}

@article{epifano2023revisiting,
  title={{Revisiting the Fragility of Influence Functions}},
  author={Epifano, Jacob R and Ramachandran, Ravi P and Masino, Aaron J and Rasool, Ghulam},
  journal={Neural Networks},
  year={2023}
}

@article{bae2022if,
  title={{If Influence Functions are the Answer, Then What is the Question?}},
  author={Bae, Juhan and Ng, Nathan and Lo, Alston and Ghassemi, Marzyeh and Grosse, Roger B},
  journal={Advances in Neural Information Processing Systems},
  year={2022}
}

@inproceedings{isal,
  title={{Influence Selection for Active Learning}},
  author={Liu, Zhuoming and Ding, Hao and Zhong, Huaping and Li, Weijia and Dai, Jifeng and He, Conghui},
  booktitle={IEEE/CVF International Conference on Computer Vision},
  year={2021}
}

@article{jain2022efficient,
title={{Efficient {D}ata {S}ubset {S}election to {G}eneralize {T}raining {A}cross {M}odels: {T}ransductive and {I}nductive {N}etworks}},
author={Eeshaan Jain and Tushar Nandy and Gaurav Aggarwal and Ashish V. Tendulkar and Rishabh K Iyer and Abir De},
journal={Advances in Neural Information Processing Systems},
year={2023},
}

@inproceedings{chen2021hydra,
  title={{Hydra: Hypergradient Data Relevance Analysis for Interpreting Deep Neural Networks}},
  author={Chen, Yuanyuan and Li, Boyang and Yu, Han and Wu, Pengcheng and Miao, Chunyan},
  booktitle={AAAI Conference on Artificial Intelligence},
  year={2021}
}

@article{killamsetty2021retrieve,
  title={{Retrieve: Coreset Selection For Efficient and Robust Semi-supervised Learning}},
  author={Killamsetty, Krishnateja and Zhao, Xujiang and Chen, Feng and Iyer, Rishabh},
  journal={Advances in Neural Information Processing Systems},
  year={2021}
}

@inproceedings{kwon2023datainf,
  title={{Data{I}nf: Efficiently {E}stimating {D}ata {I}nfluence in {L}o{RA}-tuned {LLM}s and {D}iffusion {M}odels}},
  author={Kwon, Yongchan and Wu, Eric and Wu, Kevin and Zou, James},
  booktitle={International Conference on Learning Representations},
  year={2024}
}

@inproceedings{wang2024data,
  title={{Data Shapley in One Training Run}},
  author={Wang, Jiachen T and Mittal, Prateek and Song, Dawn and Jia, Ruoxi},
  booktitle={International Conference on Learning Representations},
  year={2025}
}

@article{charpiat2019input,
  title={{Input Similarity from the Neural Network Perspective}},
  author={Charpiat, Guillaume and Girard, Nicolas and Felardos, Loris and Tarabalka, Yuliya},
  journal={Advances in Neural Information Processing Systems},
  year={2019}
}

@article{feldman2020neural,
  title={{What Neural Networks Memorize and Why: Discovering the Long Tail via Influence Estimation}},
  author={Feldman, Vitaly and Zhang, Chiyuan},
  journal={Advances in Neural Information Processing Systems},
  year={2020}
}

@article{agarwal2017second,
  title={{Second-order Stochastic Optimization for Machine Learning in Linear Time}},
  author={Agarwal, Naman and Bullins, Brian and Hazan, Elad},
  journal={The Journal of Machine Learning Research},
  year={2017},
}

@article{krizhevsky2009learning,
  title={{Learning Multiple Layers of Features from Tiny Images}},
  author={Krizhevsky, Alex and Hinton, Geoffrey and others},
  year={2009},
  journal={University of Toronto}
}

@inproceedings{he2016deep,
  title={{Deep Residual Learning for Image Recognition}},
  author={He, Kaiming and Zhang, Xiangyu and Ren, Shaoqing and Sun, Jian},
  booktitle={IEEE/CVF Conference on Computer Vision and Pattern Recognition},
  year={2016}
}

@inproceedings{bejan2023make,
  title={{Make Every Example Count: On the Stability and Utility of Self-Influence for Learning from Noisy NLP Datasets}},
  author={Bejan, Irina and Sokolov, Artem and Filippova, Katja},
  booktitle={Conference on Empirical Methods in Natural Language Processing},
  year={2023}
}

@inproceedings{deng2009imagenet,
  title={{Imagenet: A Large-scale Hierarchical Image Database}},
  author={Deng, Jia and Dong, Wei and Socher, Richard and Li, Li-Jia and Li, Kai and Fei-Fei, Li},
  booktitle={IEEE Conference on Computer Vision and Pattern Recognition},
  year={2009}
}

@article{grosse2023studying,
  title={{Studying Large Language Model Generalization with Influence Functions}},
  author={Grosse, Roger and Bae, Juhan and Anil, Cem and Elhage, Nelson and Tamkin, Alex and Tajdini, Amirhossein and Steiner, Benoit and Li, Dustin and Durmus, Esin and Perez, Ethan and others},
  journal={arXiv preprint arXiv:2308.03296},
  year={2023}
}

@inproceedings{kong2021resolving,
  title={{Resolving Training Biases via Influence-based Data Relabeling}},
  author={Kong, Shuming and Shen, Yanyan and Huang, Linpeng},
  booktitle={International Conference on Learning Representations},
  year={2021}
}

@article{tan2024data,
  title={{Data Pruning via Moving-one-Sample-out}},
  author={Tan, Haoru and Wu, Sitong and Du, Fei and Chen, Yukang and Wang, Zhibin and Wang, Fan and Qi, Xiaojuan},
  journal={Advances in Neural Information Processing Systems},
  year={2024}
}

@article{lyu2023deeper,
  title={{Deeper Understanding of Black-box Predictions via Generalized Influence Functions}},
  author={Lyu, Hyeonsu and Jang, Jonggyu and Ryu, Sehyun and Yang, Hyun Jong},
  journal={arXiv preprint arXiv:2312.05586},
  year={2023}
}

@article{dai2023training,
  title={{Training Data Attribution for Diffusion Models}},
  author={Dai, Zheng and Gifford, David K},
  journal={arXiv preprint arXiv:2306.02174},
  year={2023}
}

@inproceedings{thakkar2023self,
  title={{Self-Influence Guided Data Reweighting for Language Model Pre-training}},
  author={Thakkar, Megh and Bolukbasi, Tolga and Ganapathy, Sriram and Vashishth, Shikhar and Chandar, Sarath and Talukdar, Partha},
  booktitle={Conference on Empirical Methods in Natural Language Processing},
  year={2023}
}

@article{schioppa2024theoretical,
  title={{Theoretical and Practical Perspectives on What Influence Functions Do}},
  author={Schioppa, Andrea and Filippova, Katja and Titov, Ivan and Zablotskaia, Polina},
  journal={Advances in Neural Information Processing Systems},
  year={2024}
}

@inproceedings{richardson2023add,
  title={{Add-Remove-or-Relabel: Practitioner-Friendly Bias Mitigation via Influential Fairness}},
  author={Richardson, Brianna and Sattigeri, Prasanna and Wei, Dennis and Ramamurthy, Karthikeyan Natesan and Varshney, Kush and Dhurandhar, Amit and Gilbert, Juan E},
  booktitle={ACM Conference on Fairness, Accountability, and Transparency},
  year={2023}
}

@article{kim2024gex,
  title={{GEX: A Flexible Method for Approximating Influence via Geometric Ensemble}},
  author={Kim, SungYub and Kim, Kyungsu and Yang, Eunho},
  journal={Advances in Neural Information Processing Systems},
  year={2024}
}

@inproceedings{fair_pareto,
title	= {{What is Fair? Exploring Pareto-Efficiency for Fairness Constraint Classifiers}},
author	= {Alyssa Whitlock Lees and Ananth Balashankar and Chris Welty and Lakshminarayanan Subramanian},
year	= {2019},
booktitle	= {arXiv preprint arXiv:1910.14120}
}

@inproceedings{martinez2020minimax,
  title={{Minimax Pareto Fairness: A Multi Objective Perspective}},
  author={Martinez, Natalia and Bertran, Martin and Sapiro, Guillermo},
  booktitle={International Conference on Machine Learning},
  year={2020}
}

@incollection{lotov2008visualizing,
  title={{Visualizing the Pareto Frontier}},
  author={Lotov, Alexander V and Miettinen, Kaisa},
  booktitle={Multiobjective optimization: interactive and evolutionary approaches},
  pages={213--243},
  year={2008},
  publisher={Springer}
}

@inproceedings{wenxiao_category,
  title={{Category-Aware Active Domain Adaptation}},
  author={Xiao, Wenxiao and Gu, Jiuxiang and Liu, Hongfu},
  booktitle={International Conference on Machine Learning},
  year={2024}
}

@InProceedings{pmlr-v15-coates11a,
  title = 	 {{An Analysis of Single-Layer Networks in Unsupervised Feature Learning}},
  author = 	 {Coates, Adam and Ng, Andrew and Lee, Honglak},
  booktitle = 	 {International Conference on Artificial Intelligence and Statistics},
  year = 	 {2011}
}

@inproceedings{saravia2018carer,
  title={{CARER: Contextualized Affect Representations for Emotion Recognition}},
  author={Saravia, Elvis and Liu, Hsien-Chi Toby and Huang, Yen-Hao and Wu, Junlin and Chen, Yi-Shin},
  booktitle={Conference on Empirical Methods in Natural Language Processing},
  year={2018}
}

@article{zhang2015character,
  title={Character-level Convolutional Networks for Text Classification},
  author={Zhang, Xiang and Zhao, Junbo and LeCun, Yann},
  journal={Advances in neural information processing systems},
  volume={28},
  year={2015}
}

@article{forrest1996genetic,
  title={{Genetic Algorithms}},
  author={Forrest, Stephanie},
  journal={ACM Computing Surveys},
  year={1996}
}

@article{NEURIPS2023_960573a3,
 author = {Edelman, Benjamin and Goel, Surbhi and Kakade, Sham and Malach, Eran and Zhang, Cyril},
 journal = {Advances in Neural Information Processing Systems},
 title = {{Pareto Frontiers in Deep Feature Learning: Data, Compute, Width, and Luck}},
 year = {2023}
}

@article{NEURIPS2019_685bfde0,
 author = {Lin, Xi and Zhen, Hui-Ling and Li, Zhenhua and Zhang, Qing-Fu and Kwong, Sam},
 journal = {Advances in Neural Information Processing Systems},
 title = {{Pareto Multi-Task Learning}},
 year = {2019}
}

@article{JMLR:v20:18-598,
  author  = {Thomas Elsken and Jan Hendrik Metzen and Frank Hutter},
  title   = {{Neural Architecture Search: A Survey}},
  journal = {Journal of Machine Learning Research},
  year    = {2019},
  volume  = {20},
  number  = {55},
  pages   = {1--21}
}

@inproceedings{10.5555/3709347.3743813,
author = {R\"{o}pke, Willem and Reymond, Mathieu and Mannion, Patrick and Roijers, Diederik M and Now\'{e}, Ann and R\u{a}dulescu, Roxana},
title = {{Divide and Conquer: Provably Unveiling the Pareto Front with Multi-Objective Reinforcement Learning}},
year = {2025},
booktitle = {International Conference on Autonomous Agents and Multiagent Systems}
}

@article{NEURIPS2023_32285dd1,
 author = {Cai, Xin-Qiang and Zhang, Pushi and Zhao, Li and Bian, Jiang and Sugiyama, Masashi and Llorens, Ashley},
 journal = {Advances in Neural Information Processing Systems},
 title = {{Distributional Pareto-Optimal Multi-Objective Reinforcement Learning}},
 year = {2023}
}

@article{DBLP:journals/corr/abs-1810-04805,
  author    = {Jacob Devlin and
               Ming{-}Wei Chang and
               Kenton Lee and
               Kristina Toutanova},
  title     = {{BERT:} Pre-training of Deep Bidirectional Transformers for Language
               Understanding},
  journal   = {CoRR},
  volume    = {abs/1810.04805},
  year      = {2018},
  archivePrefix = {arXiv},
  eprint    = {1810.04805},
  bibsource = {dblp computer science bibliography, https://dblp.org}
}

@inproceedings{wu2022tinyvit,
  title={{Tinyvit: Fast Pretraining Distillation for Small Vision Transformers}},
  author={Wu, Kan and Zhang, Jinnian and Peng, Houwen and Liu, Mengchen and Xiao, Bin and Fu, Jianlong and Yuan, Lu},
  booktitle={European Conference on Computer Vision},
  year={2022}
}
\bibliographystyle{unsrtnat}

\clearpage
\appendix
\section*{Appendix}

\section{Experiments on Synthetic Datasets}\label{app:syn}
\begin{figure*}[h]
    \centering
    \includegraphics[width=0.825\linewidth]{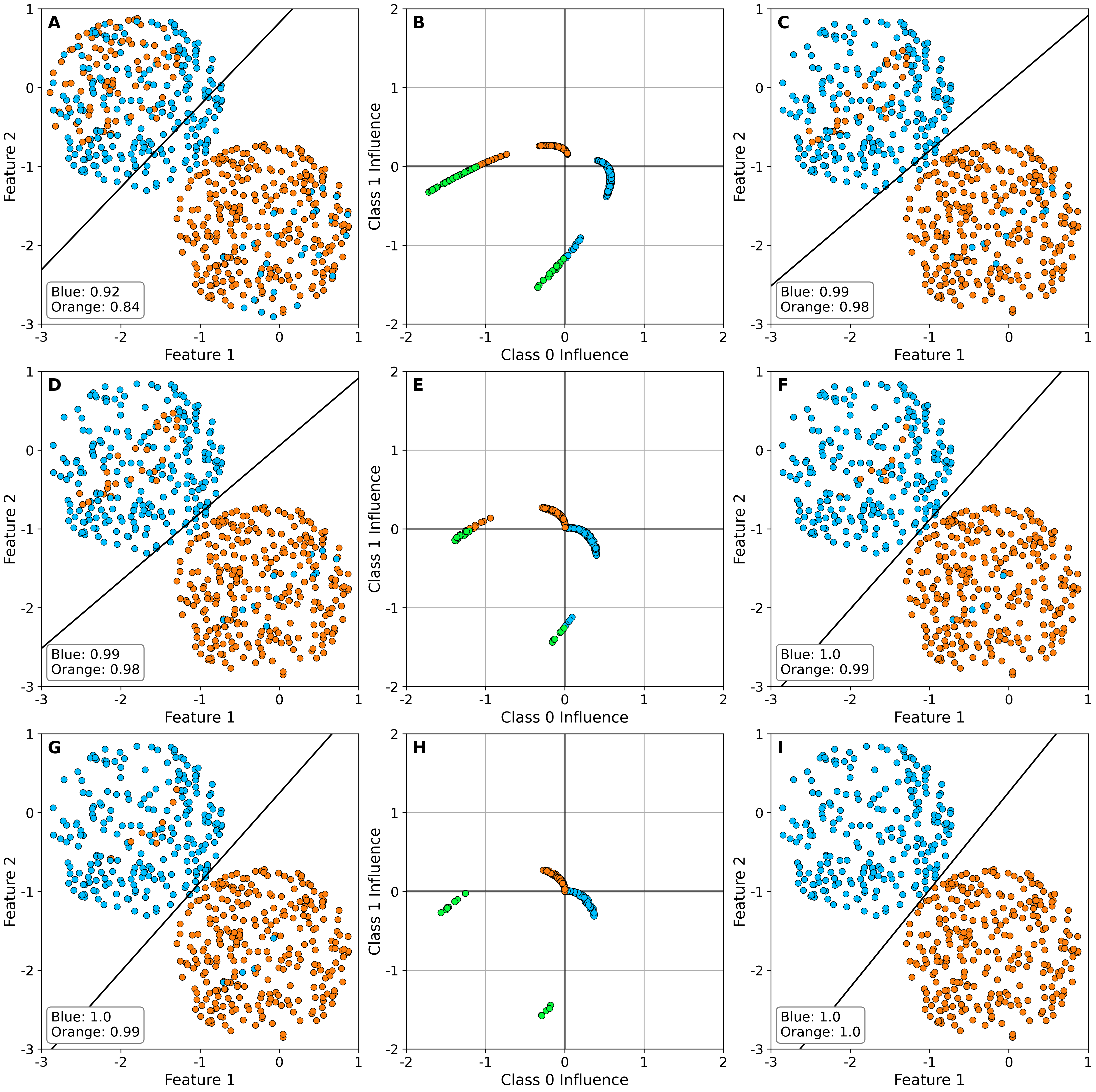}\vspace{-4mm}
    \caption{Category-wise influence function for sample removal. The dataset used in this demonstration is identical to that in the top row of Figure \ref{fig:toy}. Each row displays the state of the dataset, the changes that our method will make, and the result, of the dataset trimming procedure. Subfigures A, D, and G display the state of the training dataset. The color of each point indicates its label. The linear decision boundary is drawn, and its accuracy  across both classes is shown in the legend. Subfigures B, E, and H show the category-wise influence score of each training point. Training data points in green are indicated by the score to be detrimental to model performance on both classes. These points will be removed by the improvement procedure. Subfigures C, F, and H show the training dataset after removal. The resulting decision boundary and its accuracy is also indicated. Note how the noise furthest from the decision boundary is removed first, since these have the largest effect on the decision boundary. Additionally, through each iteration, the accuracy of the linear model is improved when using the trimmed dataset. After three iterations, all noise has been removed and the model is at the performance ceiling.}
    \label{fig:toy2}
\end{figure*}

\looseness-1 As we have discussed the technique of identifying and removing noisy training samples, specifically those exhibiting negative influence scores for both groups, we iteratively apply this procedure to the same dataset. The results are presented in Fig.~\ref{fig:toy2}. The dataset and model configuration are identical to those used in Subfigures \textbf{ABC} of Fig.~\ref{fig:toy}. In each iteration, we: (1) train a logistic regression model on the training set, (2) compute category-aware influence scores for all training samples, (3) remove samples with negative scores, and (4) retrain the model on the refined dataset. As illustrated in the figure, repeated application of this procedure progressively moves the model closer to the performance frontier.

\section{Experiments on Real-World Datasets}\label{app:real}
\begin{figure*}[h]
    \centering
    \includegraphics[width=0.98\linewidth]{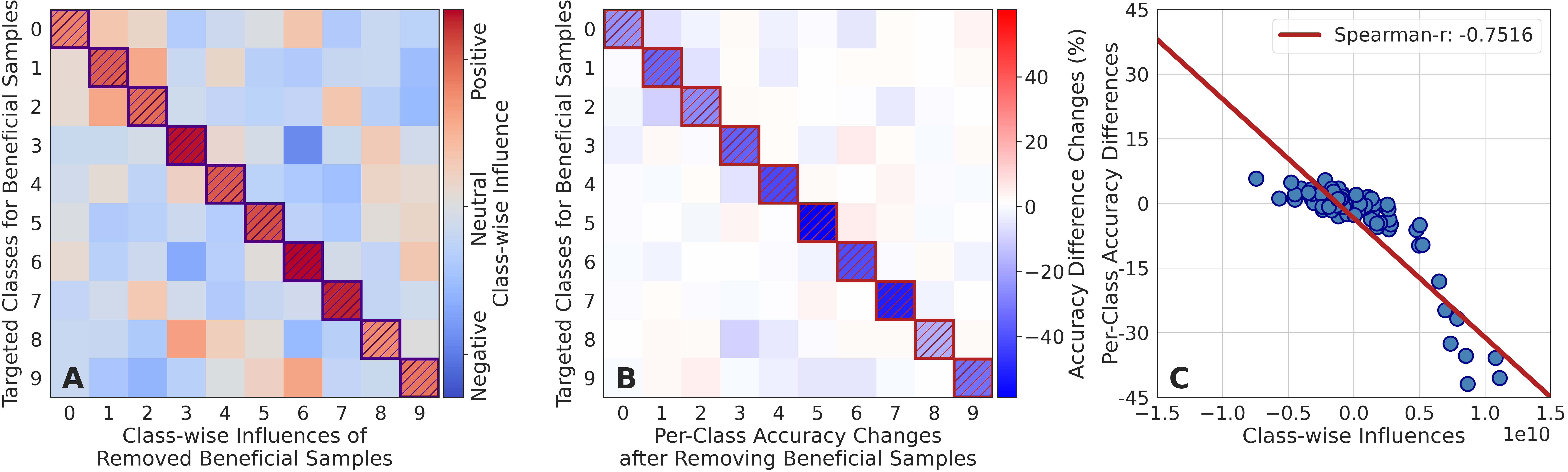}
    \includegraphics[width=0.98\linewidth]{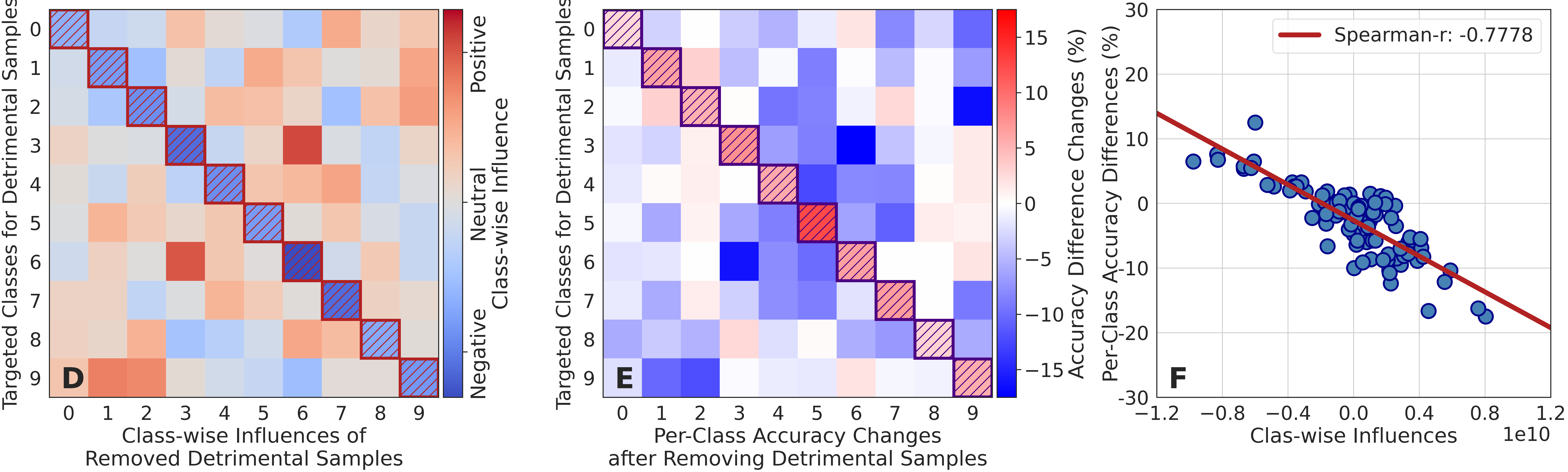}\vspace{2mm}
    \caption{Category-wise influence on \textit{STL-10}~\citep{pmlr-v15-coates11a} image dataset. Subfigures \textbf{(A, D)} denote predicted category-wise influence and \textbf{(B, E)} actual accuracy changes with beneficial/detrimental sample removal. Subfigures \textbf{(C, F)} denote scatter plots between predicted influence and actual performance shifts.}
    \label{fig:stl10}
\end{figure*}

\begin{figure*}[h]
    \centering
    \includegraphics[width=0.98\linewidth]{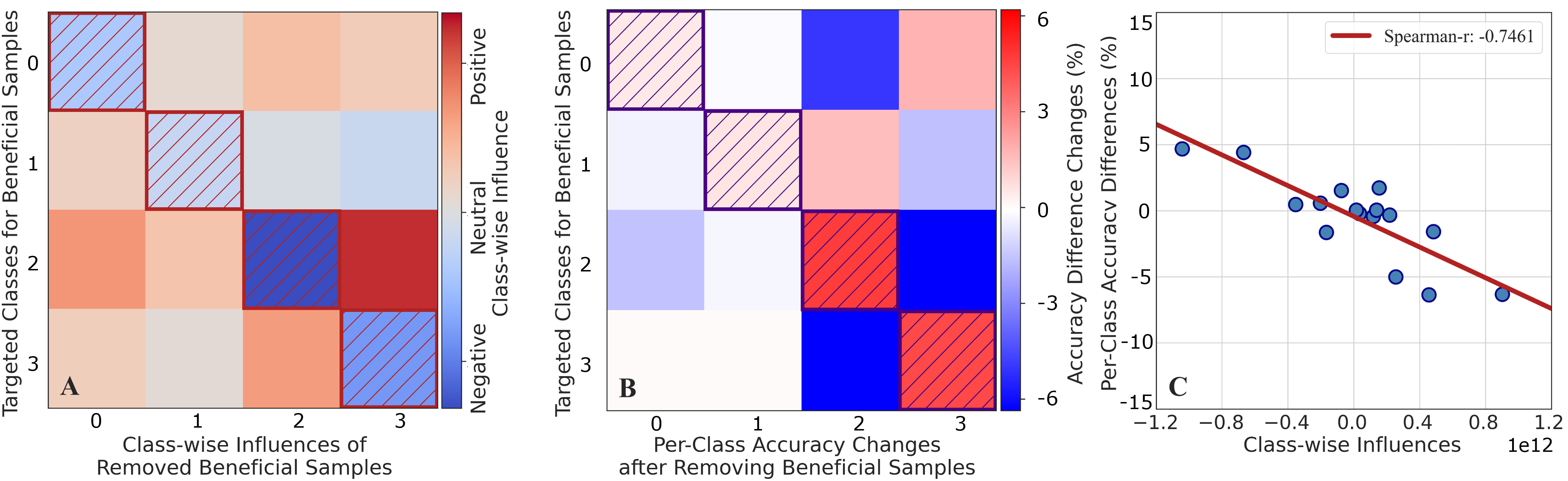}
    \includegraphics[width=0.98\linewidth]{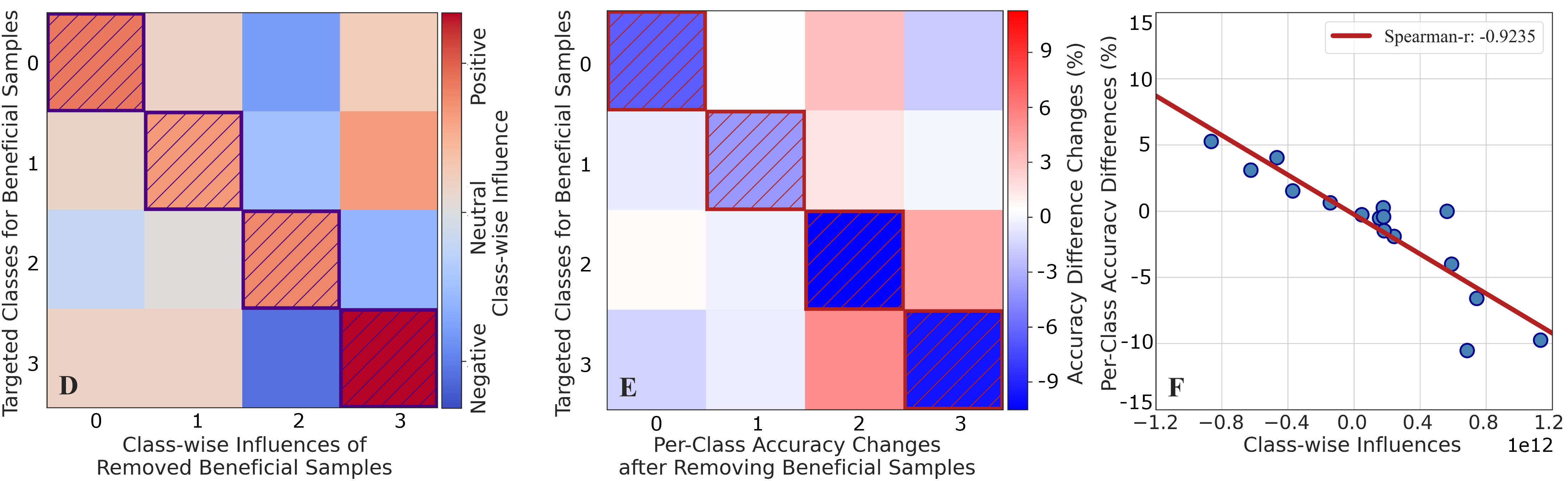}\vspace{2mm}
    \caption{Category-wise influence on \textit{AG\_News}~\citep{zhang2015character} text dataset. This figure follows the same setup and caption format as Fig.~\ref{fig:stl10}.}
    \label{fig:agnews}
\end{figure*}

Category-wise influences on \textit{STL-10} and \textit{AG\_News} are shown in Fig.~\ref{fig:stl10} and~\ref{fig:agnews}.

\vfill

\end{document}